\documentclass[journal]{IEEEtran}

\ifCLASSINFOpdf
\else
\fi
\usepackage{graphicx}
\usepackage{amsmath,amssymb} % define this before the line numbering.
\usepackage{subfigure}
\usepackage{algorithm}
\usepackage{algorithmic}
\usepackage{color}
\usepackage{footmisc}
\usepackage{url}

\begin{document}

\title{Deep Metric Learning with Density Adaptivity} % Replace with your title

\author{Yehao~Li,
        Ting~Yao,~\IEEEmembership{Member,~IEEE,}
        Yingwei~Pan,
        Hongyang~Chao,~\IEEEmembership{Member,~IEEE,}\\
        and~Tao~Mei,~\IEEEmembership{Fellow,~IEEE}
\thanks{T. Yao is the corresponding author.}
\thanks{This work is partially supported by NSF of China under Grant 61672548, U1611461, 61173081, and the Guangzhou Science and Technology Program, China, under Grant 201510010165.}
\thanks{Y. Li and H. Chao are with Sun Yat-Sen University, Guangzhou, China (e-mail: yehaoli.sysu@gmail.com; isschhy@mail.sysu.edu.cn).}
\thanks{T. Yao, Y. Pan and T. Mei are with JD AI Research, Beijing, China (e-mail: tingyao.ustc@gmail.com; panyw.ustc@gmail.com; tmei@jd.com).}}

%% The paper headers

\maketitle

\begin{abstract}
The problem of distance metric learning is mostly considered from the perspective of learning an embedding space, where the distances between pairs of examples are in correspondence with a similarity metric. With the rise and success of Convolutional Neural Networks (CNN), deep metric learning (DML) involves training a network to learn a nonlinear transformation to the embedding space. Existing DML approaches often express the supervision through maximizing inter-class distance and minimizing intra-class variation. However, the results can suffer from overfitting problem, especially when the training examples of each class are embedded together tightly and the density of each class is very high. In this paper, we integrate density, i.e., the measure of data concentration in the representation, into the optimization of DML frameworks to adaptively balance inter-class similarity and intra-class variation by training the architecture in an end-to-end manner. Technically, the knowledge of density is employed as a regularizer, which is pluggable to any DML architecture with different objective functions such as contrastive loss, N-pair loss and triplet loss. Extensive experiments on three public datasets consistently demonstrate clear improvements by amending three types of embedding with the density adaptivity. More remarkably, our proposal increases Recall@1 from 67.95\% to 77.62\%, from 52.01\% to 55.64\% and from 68.20\% to 70.56\% on Cars196, CUB-200-2011 and Stanford Online Products dataset, respectively.
\end{abstract}


\section{Introduction}
\IEEEPARstart{L}{earning} to assess the distance between the pairs of examples or learning a good metric is crucial in machine learning and real-world multimedia applications. One typical direction to define and learn metrics that reflect succinct characteristics of the data is from the viewpoint of classification, where a clear supervised objective, i.e., classification error, is available and could be optimized for. However, there is no guarantee that classification approaches could learn good and general metrics for any tasks, particularly when the data distribution at test time is quite different not to mention that some test examples are even from previously unseen classes. More importantly, the extreme case with enormous number of classes and only a few labeled examples per class practically stymies the direct classification. Distance metric learning, in contrast, aims at learning a transformation to an embedding space, which is regarded as a full metric over the input space by exploring not only semantic information of each example in the training set but also their intra-class and inter-class structures. As such, the learnt metric generalizes more easily.

\begin{figure*}[!tb]
   \centering
   \subfigure[Contrastive Loss]{
     \label{fig:fig1:a}
     \includegraphics[width=0.21\textwidth]{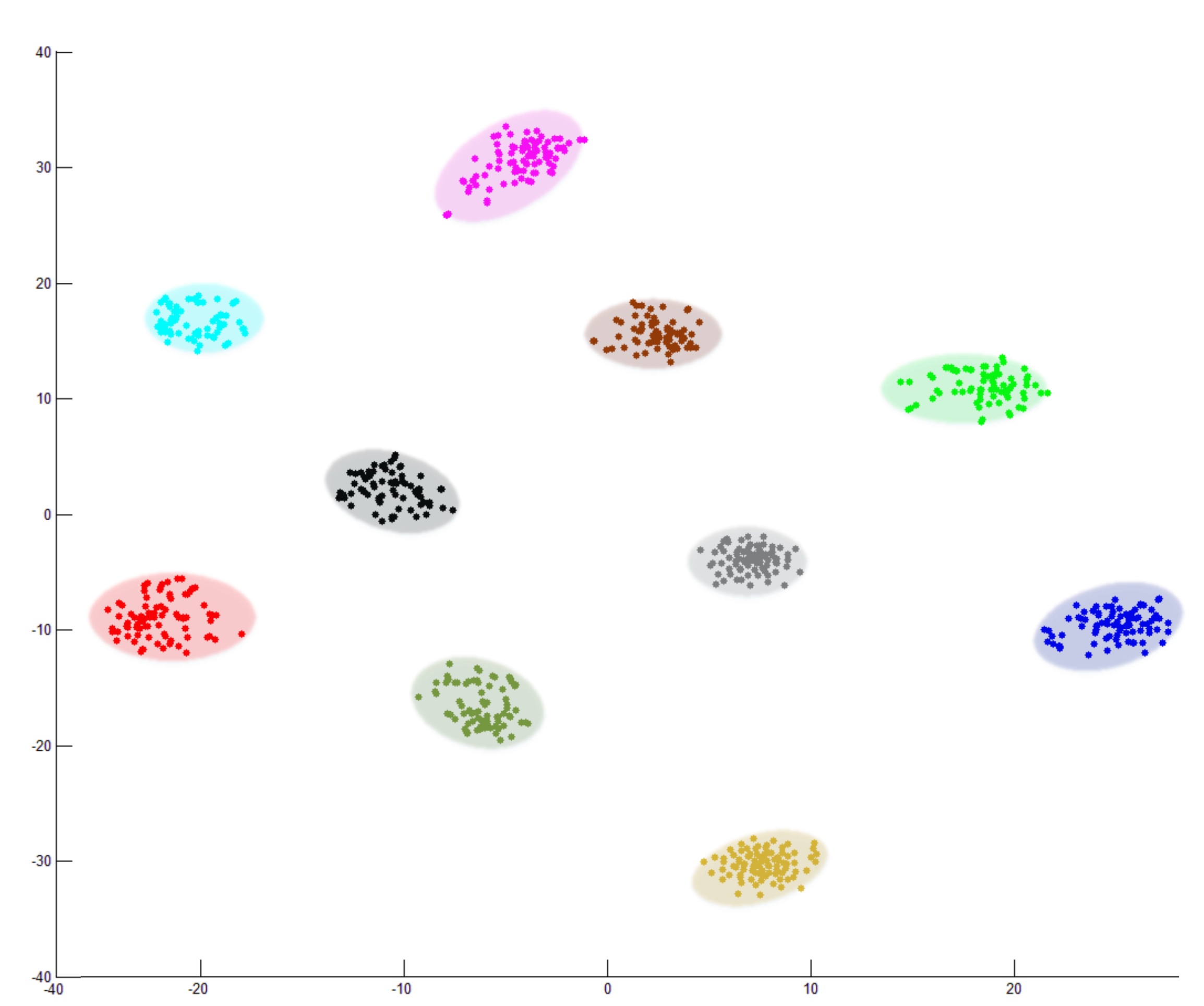}}
   \subfigure[N-pair Loss]{
     \label{fig:fig1:b}
     \includegraphics[width=0.21\textwidth]{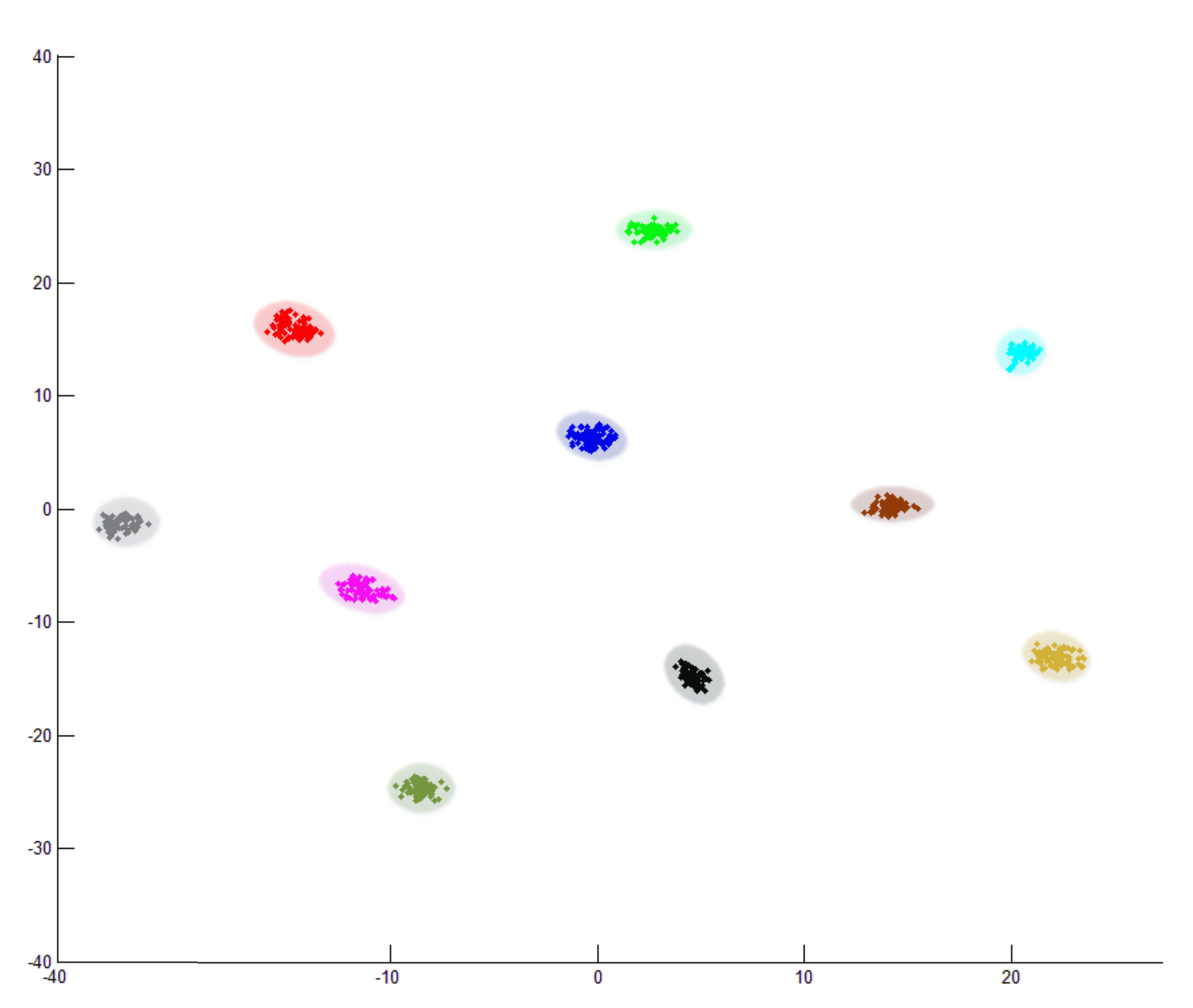}}
   \subfigure[Triplet Loss]{
     \label{fig:fig1:c}
     \includegraphics[width=0.21\textwidth]{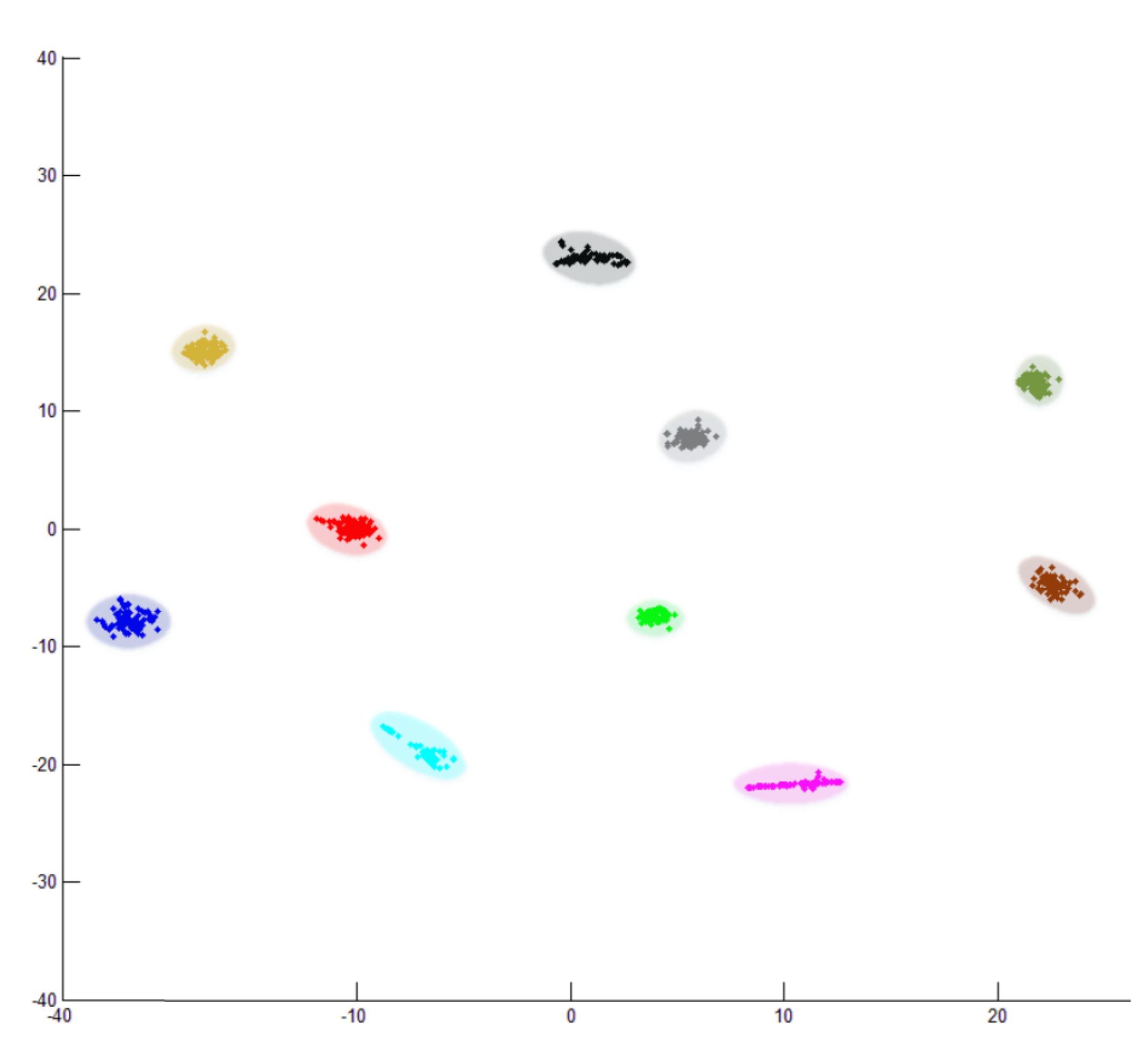}}
   \subfigure[Recall@1 Comparison]{
     \label{fig:fig1:d}
     \includegraphics[width=0.32\textwidth]{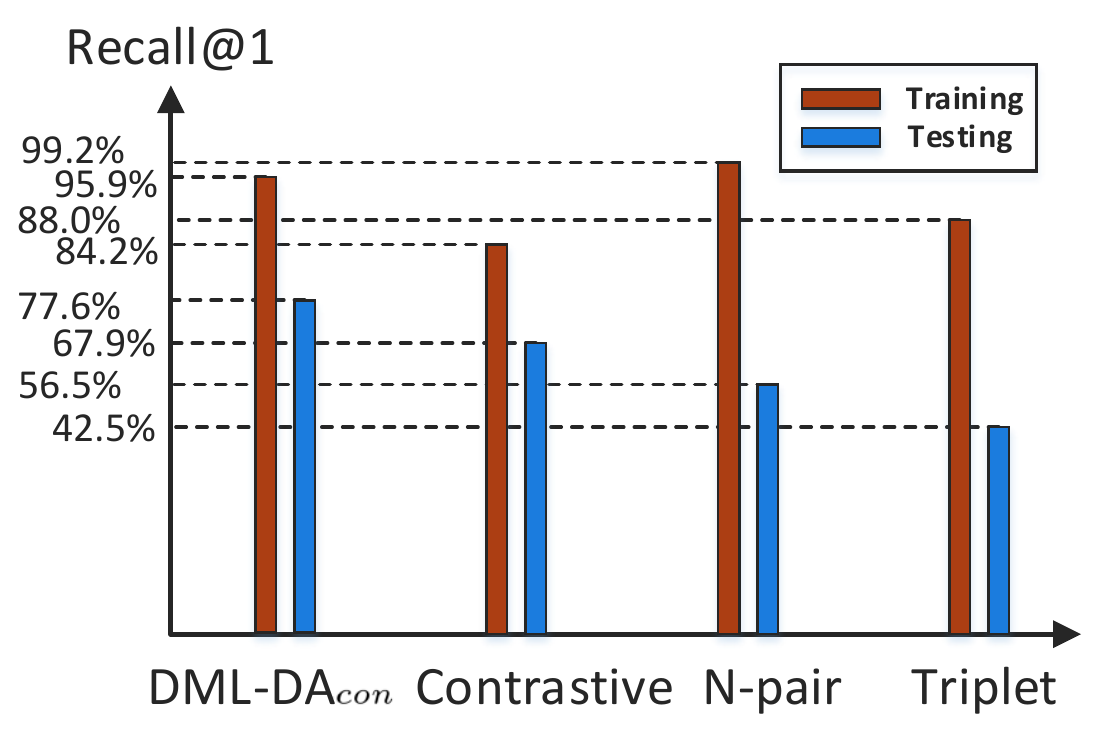}}
   \vspace{-0.05in}
   \caption{(a)---(c): Image representation embedding visualizations of ten randomly selected training classes from Cars196 dataset by using t-SNE \cite{maaten:JMLR08}. Each image is visualized as one point and colors denote different classes. The embedding space is learnt by a standard DML architecture with contrastive loss, N-pair loss and triplet loss, respectively. (d): Recall@1 performance on training \& testing set by optimizing different losses and by regularizing contrastive embedding with our density adaptivity (DML-DA$_{con}$).}
   \vspace{-0.10in}
   \label{fig:fig1}
\end{figure*}

The recent attempts on metric learning are inspired by the advances of using deep learning and learn an embedding representation of the data through neural networks. Deep metric learning (DML) has demonstrated high capability in a wide range of multimedia tasks, e.g., visual product search \cite{Huang:NIPS16,Song:CVPR16,Ustinova:NIPS16}, image retrieval \cite{li2015weakly,Rippel:ICLR16,Sohn:NIPS16,li2019learning,yao2015learning}, clustering \cite{Hershey:ICASSP16}, zero-shot image classification \cite{cui2018general,Zhang:CVPR16}, highlight detection \cite{kim2018exploiting,yao2016highlight}, face recognition \cite{Schroff:CVPR15,Sun:NIPS14} and person re-identification \cite{bai2018group,ma2014person}. The basic objective of the learning process is to preserve similar examples close in proximity and make dissimilar examples far apart from each other in the embedding space. To achieve this objective, a broad variety of losses, e.g., contrastive loss \cite{bell15productnet,Hadsell:CVPR06}, N-pair loss \cite{Sohn:NIPS16} and triplet loss \cite{Schroff:CVPR15,Weinberger:NIPS06}, are devised to explore the relationship between pairs or triplets of examples. Nonetheless, there is no clear picture of how to control the generalization error, i.e., difference between ``training error" and ``test error," when capitalizing on these losses. Take Cars196 dataset \cite{Krause:ICCV13} as an example, a standard DML architecture with N-pair loss fits the training set nicely and achieves Recall@1 performance of 99.2\% but generalizes poorly on the testing set and only reaches 56.5\% Recall@1 as shown in Figure \ref{fig:fig1:d}. Similarly, the generalization error is also observed when employing contrastive loss and triplet loss. Among the three losses, utilizing contrastive loss expresses the smallest generalization error and exhibits the highest performance on the testing set. More interestingly, the embedding representations of images from each class in the training set are more concentrated by using N-pair loss and triplet loss than contrastive loss as visualized in Figure \ref{fig:fig1:a}---\ref{fig:fig1:c}. In other words, optimizing contrastive loss leads to low density of example concentration. Here density refers to the measure of data concentration in the representation. This observation motivates us to explore the fuzzy relationship between density of examples in the embedding space and generalization capability of DML.

By consolidating the idea of exploring density to mitigate overfitting, we integrate density adaptivity into metric learning as a regularizer, following the theory that some form of regularization is needed to ensure small generalization error \cite{Zhang:ICLR17}. The regularizer of density could be easily plugged into any existing DML framework by training the whole architecture in an end-to-end fashion. We formulate the density regularizer such that it enlarges intra-class variation while the loss in DML penalizes representation distribution overlap across different classes in the embedding space. As such, the embedding representations could be sufficiently spread out to fully utilize the expressive power of the embedding space. Moreover, considering that the inherent structure of each class should be preserved before and after representation embedding, relative relationship with respect to density between different classes is further taken into account to optimize the whole architecture. Technically, the target density of each class can be viewed as an intermediate variable in our designed regularizer. It is natural to simultaneously learn the target density of each class and the neural networks by optimizing the whole architecture through the DML loss plus density regularizer. As illustrated in Figure \ref{fig:fig1:d}, contrastive embedding with our density adaptivity further decreases the generalization error and boosts up Recall@1 performance to 77.6\% on Cars196 testing set.

The main contribution of this work is the proposal of density adaptivity for addressing the issue of model generalization in the context of distance metric learning. This also leads to the elegant view of what role the density should act as in DML framework, which is a problem not yet fully understood in the literature. Through an extensive set of experiments, we demonstrate that our density adaptivity is amenable to three types of embedding with clear improvements on three different benchmarks. The remaining sections are organized as follows. Section \ref{sec:RW} describes the related works. Section \ref{sec:DMLDA} presents our approach of deep metric learning with density adaptivity, while Section \ref{sec:EX} presents the experimental results for image retrieval. Finally, Section \ref{sec:CON} concludes this paper.

\section{Related Work}\label{sec:RW}
The research on deep metric learning has mainly proceeded along two basic types of embedding, i.e., contrastive embedding and triplet embedding. The spirit of contrastive embedding is to make each positive pair from the same class in close proximity and meanwhile push the two samples in each negative pair to become far apart from each other. That is to pursue a discriminative embedding space with pairwise supervision. \cite{Sun:NIPS14} is one of the early works to capitalize on contrastive embedding for deep metric learning in face verification task. The method learns embedding space through two identical sub-networks with the input pairs of samples. Next, an amount of subsequent works are presented to leverage contrastive embedding in several practical applications, e.g., person re-identification \cite{ahmed2015improved,li2014deepreid} and image retrieval \cite{liu2016deep,xia2014supervised}. As an extension of contrastive embedding, triplet embedding \cite{hoffer2014deep,wan2014deep,pan2015semi,qiu2017deep,pan2014click} is another dimension of DML approaches by learning embedding function with triplet/ranking supervision over the set of ranking triplets. For each input triplet consisting of one query sample, one positive sample from the same class and another negative sample from different classes, the training procedure can be interpreted as the preservation of relative similarity relations like ``for the query sample, it should be more similar to positive sample than to negative sample."

Despite the promising success of both contrastive embedding and triplet embedding in aforementioned tasks, the two embeddings rely on huge amounts of pairs or triplets for training, resulting in slow convergence and even local optimization. This is partially due to the fact that existing methods often construct each mini-batch with randomly sampled pairs or triplets and the loss functions are measured independently over individual pairs or triplets without any interaction among them. To alleviate the problem, a practical trick, i.e., hard sample mining \cite{cui2016fine,simo2015discriminative,wang2014learning,pan2016learning}, is commonly leveraged to accelerate convergence with the hard pairs or triplets selected from each mini-batch. In particular, \cite{wang2014learning} devises an effective hard triplet sampling strategy by selecting more positive images with higher relevance scores and hard in-class negative images with less relevance scores. In another work \cite{simo2015discriminative}, the idea of hard mining is incorporated into contrastive embedding by gradually searching hard negative samples for training.

Recently, a variety of works design new loss functions for training, pursuing more effective DML. For example, \cite{bell15productnet,zhang2016embedding} present a simple yet effective method by combining deep metric learning with classification constraint in a multi-task learning framework. \cite{Sohn:NIPS16} develops N-pair embedding which improves triplet embedding by pushing away multiple negative samples simultaneously within a mini-batch. Such design of N-pair embedding constructs each batch with N pairs of samples, leading to more efficient convergence in training stage. Song \emph{et al.} define a structured prediction objective for DML by lifting the examples within a batch into a dense pairwise matrix in \cite{Song:CVPR16}. Later in \cite{song2016learnable}, another structured prediction-based method is designed to directly optimize the deep neural network with a clustering quality metric. Ustinova \emph{et al.} propose a new Histogram loss \cite{Ustinova:NIPS16} to train the deep embeddings through making the distribution of similarities of positive and negative pairs less overlapped. Huang \emph{et al.} introduce a Position-Dependent Deep Metric (PDDM) unit \cite{Huang:NIPS16} which is capable of learning a similarity metric adaptive to local feature structure. Most recently, in \cite{yuan2017hard}, a Hard-Aware Deeply Cascaded embedding (HDC) is devised to handle samples of different hard level with sub-networks of different depths in a cascaded manner. \cite{Zhang:ICCV17} presents a global orthogonal regularizer to improve DML with pairwise and triplet losses by making two randomly sampled non-matching embedding representations close to orthogonal.

In the literature, there have been few works, being proposed for exploiting the adaptation of density in deep metric learning. \cite{Rippel:ICLR16} is by arbitrarily splitting the distributions of classes in representation space to pursue local discrimination. Technically, the method maintains a number of clusters for each class and adaptively embraces intra-class variation and inter-class similarity by minimizing intra-cluster distances. As such, high density of data concentration is encouraged in each cluster. Instead, our work adapts data concentration through maximizing the feature spread or seeking to low density of feature distribution for each class, while guaranteeing all the classes separable. As a result, the expressive capability of the representation space could be fully endowed to enhance model generalization, making our model potentially more effective and robust. Moreover, relative relationship with respect to density between different classes is further taken into account to optimize DML architecture in our framework.

\begin{figure*}[!tb]
    \centering {\includegraphics[width=1.0\textwidth]{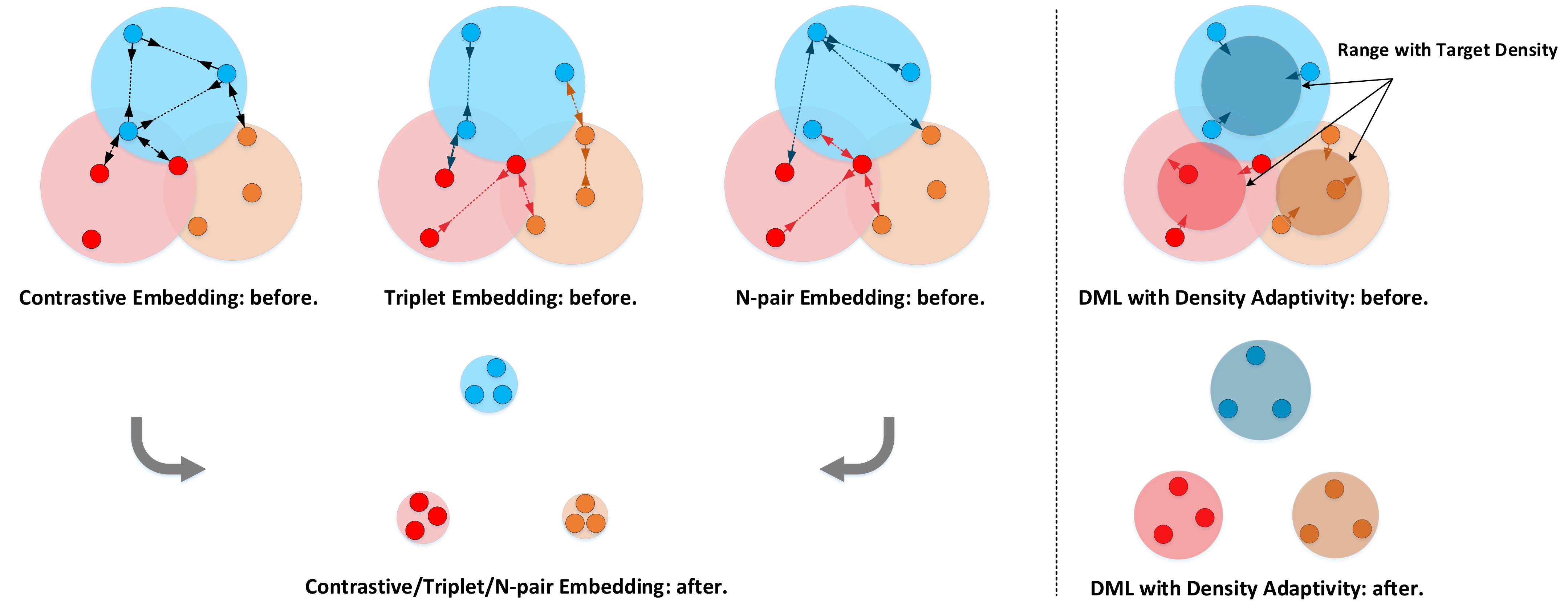}}
      \vspace{-0.15in}
    \caption{The intuition behind existing DML models (e.g., Contrastive Embedding \cite{bell15productnet}, Triplet Embedding \cite{Weinberger:NIPS06}, N-pair Embedding \cite{Sohn:NIPS16}) and our proposed DML with Density Adaptivity. The three DML models are all optimized by maximizing inter-class distance and minimizing intra-class variation, often resulting in overfitting problem as the examples of each class are enforced to be concentrated tightly, i.e., the density of each class is very high. In contrast, for DML with our proposed density regularizer, at each iteration the density of each class is estimated and adapted towards a target of low density which encourages to enlarge intra-class variation while guaranteeing all the classes seperable. Meanwhile, the objective in DML penalizes representation distribution overlap across different classes. Such balance between inter-class similarity and intra-class variation leads to better generalization capability of DML model.}
    \label{fig:5}
    \vspace{-0.15in}
\end{figure*}

\section{Deep Metric Learning with Density Adaptivity}\label{sec:DMLDA}
Our proposed Deep Metric Learning with Density Adaptivity (DML-DA) approach is to build an embedding space in which the feature representations of images could be encoded with semantic supervision over pairs or triplets of examples, under the umbrella of density adaptivity for each class. The training of DML-DA is performed by simultaneously maximizing inter-class distance and minimizing intra-class variation, and dynamically adapting the density of each class to further regularize intra-class variation, targeting for better model generalization. Therefore, the objective function of DML-DA consists of two components, i.e., the standard DML loss over pairs or triplets of examples and the proposed density regularizer. In the following, we will first recall basic methods of DML, followed by presenting how to estimate and adapt the density of each class as a regularizer. Then, we formulate the joint objective function of DML with density adaptivity and present the optimization strategy in one deep learning framework. Specifically, a DML loss layer with density regularizer is elaborately devised to optimize the whole~architecture.

\subsection{Deep Metric Learning}\label{ssec:DML}
Suppose we have a training set with $N$ examples of image-label pairs $\mathcal{S} = \{(x_i,y_i)\}^N_{i=1}$ belonging to $C$ classes, where ${y_i} \in \left\{ {1,2,...,C} \right\}$ is the class label of image $x_i$. With the standard setting of deep metric learning, the target is to learn an embedding function $f\left( {x_i;\theta } \right):x_i \to {\mathbb{R}}^d$ for transforming each input image into a $d$-dimensional embedding space through a deep architecture, where $\theta$ represents the learnable parameters of the deep neural networks. Note that length-normalization is performed on the top of the deep architecture, making all the embedded representations $\left\{ {f\left( {x_i;\theta } \right)} \right\}$ $L_2$-normalized. Given two images $x_i$ and $x_j$, the most natural way to measure the relations between them is to calculate the Euclidean distance in the embedding space as
\begin{equation}\label{Eq:Eq1}
\mathcal{D} \left({x_i,x_j}\right)= \left\| {f\left( {{x_i};\theta } \right) - f\left( {{x_j};\theta } \right)} \right\|_2^2.
\end{equation}
After taking such Euclidean distance as a similarity metric, the concrete task for DML is to learn the discriminative embedding representation by preserving the semantic relationship underlying in pairs \cite{Bromley:NIPS94,Chopra:CVPR05}, or triplets \cite{Schroff:CVPR15,Weinberger:NIPS06} or even more critical relationships (e.g., N-pair loss \cite{Sohn:NIPS16}).

\textbf{Contrastive Embedding.} Contrastive embedding is the most popular DML method which aims to generate embedding representations to satisfy the pairwise supervision, i.e., making the distance between a positive pair of examples from the same class minimized while maximized on a negative pair from different classes. Concretely, the corresponding contrastive loss function is defined as
\begin{equation}\label{Eq:Eq2}\scriptsize
{\mathcal{L}_{pair}} = \sum\limits_{\left( {{x_i},{x_j}} \right) \in \mathcal{M}} {\mathcal{D}\left( {{x_i},{x_j}} \right)}  + \sum\limits_{\left( {{x_i},{x_j}} \right) \in \mathcal{C}} {\max \left( {0,m_p - \mathcal{D}\left( {{x_i},{x_j}} \right)} \right)},
\end{equation}
where $m_p$ is the functional margin in the hinge function. $\mathcal{M}$ and $\mathcal{C}$ denotes the set of positive pairs and negative pairs, respectively.

\textbf{Triplet Embedding.} Different from pairwise embedding which only considers the absolute values of distances between positive and negative pairs, triplet embedding focuses more on the relative distance ordering among triplets $\{(x_i,x^+_j,x^-_k)\}$, where $(x_i,x^+_j)$ denotes positive pair and $(x_i,x^-_k)$ is negative pair. The assumption is that the distance between negative pair $(x_i,x^-_k)$ should be larger than that of positive pair $(x_i,x^+_j)$. Hence, the triplet loss function is measured by
\begin{equation}\label{Eq:Eq3}\scriptsize
{\mathcal{L}_{triplet}} = \sum\limits_{\left( {{x_i},x_j^ + ,x_k^ - } \right) \in \mathcal{T}} {\max \left( {0,{m_t} + \mathcal{D}\left( {{x_i},x_j^ + } \right) - \mathcal{D}\left( {{x_i},x_k^ - } \right)} \right)},
\end{equation}
where $m_t$ is the enforced margin in the hinge function and $\mathcal{T}$ is the triplet set generated on $\mathcal{S}$.

\textbf{N-pair Embedding.} N-pair embedding is one recent DML model which generalizes triplet loss by encouraging the joint distance comparison among more than one negative pair. Specifically, given a $(C+1)$-tuplet of training samples $\{x_i,x_j^+,x^-_{k^{(1)}},...,x^-_{k^{(C-1)}}\}$ where $x_i$ is the anchor point, $x^+_j$ is a positive sample sharing the same label with $x_i$ and $\{x^-_{k^{(c)}}\}^{C-1}_{c=1}$ are the $(C-1)$ negative samples from the rest $(C-1)$ different classes, the N-pair loss function is then formulated as

\begin{equation}\label{Eq:Eq4}\scriptsize
\begin{split}
&{\mathcal{L}_{N - pair}} = \sum\limits_{\{ {x_i},x_j^ + ,\{ x_{{k^{(c)}}}^ - \} _{c = 1}^{C - 1}\}  \in \mathcal{N}} \log \left( {1 + \sum\limits_{c = 1}^{C - 1} {{e^{\mathcal{D}_{N}\left( {{x_i},x_j^ +,{x_{{k^{(c)}}}^ -}} \right)}}} } \right),\\
&\mathcal{D}_{N}\left( {{x_i},x_j^ +,{x_{{k^{(c)}}}^ -}} \right)={\left\{ {f{{\left( {{x_i};\theta } \right)}^\top}f\left( {x_{{k^{(c)}}}^ - ;\theta } \right)-f{{\left( {{x_i};\theta } \right)}^\top}f\left( {x_j^ + ;\theta } \right)} \right\}},
\end{split}
\end{equation}
where $\mathcal{N}$ is the $(C+1)$-tuplet set constructed over $\mathcal{S}$. Through minimizing this N-pair loss, the similarity between positive pair is enforced to be larger than all the rest $(C-1)$ negative pairs, which further enhances triplet loss in triplet embedding with more semantic supervision.

\subsection{Density Regularizer}
One of the key attributes which the recent DML methods aforementioned in Section \ref{ssec:DML} have in common is that their objectives are predominantly designed for maximizing inter-class distance and minimizing intra-class variation. Although such optimization matches the intention of encoding semantic supervision into the learnt embedding representations, it may stymie the intrinsic intra-class variation by enforcing the examples of each class to be concentrated together tightly, which often results in overfitting problem. To overcome this issue, we devise a novel regularizer for DML that encourages low density of data concentration in the learnt embedding space to achieve a better balance between inter-class distance and intra-class variation. A caricature illustrating the intuition behind the devised density regularizer is shown in Figure \ref{fig:5}.

\begin{algorithm*}[!tb]
\caption{The training of DML with density regularizer}\label{ag:ag01}
\begin{algorithmic}[1]
    \STATE
        Given a tradeoff parameter $\lambda$.
    \STATE
        \textbf{Forward Pass:}
    \STATE
        ~Fetch input batch $\tilde {\mathcal{S}}$ with $\tilde N$ sampled image-label pairs $\{(x_i,y_i)\}^{\tilde N}_{i=1}$.
    \STATE
        ~Generate the positive pairs set $\mathcal{M}$ and negative pairs set $\mathcal{C}$.
    \STATE
        ~Compute the contrastive loss over $\mathcal{M}$ and $\mathcal{C}$ via Eq. (\ref{Eq:Eq2}) and the density regularizer via Eq.(\ref{Eq:Eq8}).
    \STATE
        ~Compute overall loss output with tradeoff parameter $\lambda$.
    \STATE
        \textbf{Backward Pass:}
    \STATE
        ~Compute the overall gradient with respect to the target density of each class $\left\{ {{\alpha _c}} \right\}_{c = 1}^C$ and update the corresponding target density value.
    \STATE
        ~Compute the overall gradient with respect to input embedding representations and backward it to lower layers for updating the parameters $\theta$ of the embedding function $f\left( {x_i;\theta } \right)$.
\end{algorithmic}
\end{algorithm*}

\textbf{Density Adaptivity.}
In our context, density is a measure of data concentration in the representation space. We assume that, for the image examples belonging to the same class, high density is equivalent to the fact that all the examples are close in proximity to the corresponding class centroid. Accordingly, for class $c$, to estimate its density, one natural way is to measure the average intra-class distance between examples and the class centroid in the embedding space, which is written as
\begin{equation}\label{Eq:Eq5}\small
\begin{split}
&{\mathcal{D}_{avg}}\left( {{\mathcal{S}_c}} \right) = \frac{1}{{\left| {{\mathcal{S}_c}} \right|}}\sum\limits_{{x_i} \in {\mathcal{S}_c}} {\left\| {f\left( {{x_i};\theta } \right) - {{\mu _c}} } \right\|_2^2},\\
&{{\mu _c}}  = \frac{1}{{\left| {{\mathcal{S}_c}} \right|}}\sum\limits_{{x_i} \in {\mathcal{S}_c}} {f\left( {{x_i};\theta } \right)},
\end{split}
\end{equation}
where ${\mathcal{S}_c}$ denotes the set of samples from the same class $c$ and ${\mu _c}$ is the corresponding class centroid. Here we directly obtain the class centroid by performing mean pooling over all the samples in ${\mathcal{S}_c}$ for simplicity. The higher the density of one class, the smaller the average intra-class distance between examples belonging to this class and the class centroid of this class in the embedding space.

Based on the observations of the fuzzy relationship between density and generalization capability of DML, we propose a density regularizer to dynamically adapt the density of data concentration in the learnt embedding space for enhancing the generalization capability. The objective function of density regularizer is defined~as
\begin{equation}\label{Eq:Eq6}\small
{\mathcal{L}_{DA}} = \frac{1}{C}\sum\limits_{c = 1}^C {{{\left( {{\mathcal{D}_{avg}}\left( {{\mathcal{S}_c}} \right) - {\alpha _c}} \right)}^2}}  - \frac{1}{C}\sum\limits_{c = 1}^C {{\alpha _c}},
\end{equation}
where ${D_{avg}}\left( {{S_c}} \right)$ represents the density measurement of class $c$ in the embedding space as defined in Eq.(\ref{Eq:Eq5}). $\alpha _c$ is a newly incorporated intermediate variable which can be interpreted as the target density of class $c$ corresponding to an appropriate intra-class variation. Note that $\alpha _c$ is an intermediate which can be interpreted as the target density of class $c$. Similar to the density estimation of class $c$ in Eq.(\ref{Eq:Eq5}), $\alpha _c$ corresponds to an appropriate target intra-class variation of class $c$. The larger the value of $\alpha _c$, the lower the density of data concentration for class $c$. By minimizing this regularizer, the density of each class is enforced to be adapted towards the target density via the first term. Meanwhile, when minimizing the second term, each $\alpha _c$ is enlarged (i.e., the target intra-class variation of each class is maximized), pursuing the lower density of data concentration in each class to enhance model generalization. The rationale of our devised density regularizer is to encourage the spread-out property in a way that the regularizer adapts data concentration and maximizes the feature spread in the embedding space, while guaranteeing all the classes separable. As such, the expressive capability of the embedding space could be fully endowed. Please also note that the devised density regularizer should be jointly utilized with a basic DML model in practice, as the objective in DML is required to simultaneously prevent the intra-class variation from increasing endlessly by penalizing representation distribution overlap across different classes.

\textbf{Inter-class Density Correlations Preservation Constraint.}
Inspired by the idea of structure preservation or manifold regularization in \cite{Melacci:JMIR11}, the inter-class density correlation here is integrated into the density regularizer as a constraint to further explore the inherent density relationships between different classes. The spirit behind this constraint is that the target densities of two classes with similar inherent structures should still be similar on the embedding space. The intrinsic structure of the data in each class can be appropriately measured by the original density measurement before embedding. Specifically, our density regularizer with the constraint of inter-class density correlations is defined as
\begin{equation}\label{Eq:Eq7}\small
\begin{split}
& {\mathcal{L}_{DA}} = \frac{1}{C}\sum\limits_{c = 1}^C {{{\left( {{\mathcal{D}_{avg}}\left( {{\mathcal{S}_c}} \right) - {\alpha _c}} \right)}^2}}  - \frac{1}{C}\sum\limits_{c = 1}^C {{\alpha _c}}, \\
& s.t. ~~\frac{{{\alpha _{{c_i}}}}}{{{\alpha _{{c_j}}}}} = {\left( {\frac{{\mathcal{D}_{avg}^{\left( 0 \right)}\left( {{\mathcal{S}_{{c_i}}}} \right)}}{{\mathcal{D}_{avg}^{\left( 0 \right)}\left( {{\mathcal{S}_{{c_j}}}} \right)}}} \right)^\eta }, {c_i},{c_j} \in \left\{ {1,...,C} \right\},
\end{split}
\end{equation}
where ${\mathcal{D}_{avg}^{\left( 0 \right)}\left( {{\mathcal{S}_{{c_i}}}} \right)}$ denotes the original average intra-class distance of class $c_i$ corresponding to the original density and it is calculated based on the image representations before embedding, i.e., the output of 1,024-way $pool5/7\times7\_s1$ layer of GoogleNet \cite{Szegedy14} in our experiments. $\eta$ is utilized to control the impact of the original density and reflects what degree of the inherent density relationship between different classes is considered for measuring the density.

To make the optimization of our density regularizer easy to be solved, we relax the constraint of inter-class density correlations by appending the converted soft penalty term to the objective function and then Eq.(\ref{Eq:Eq7}) is rewritten as
\begin{equation}\label{Eq:Eq8}\scriptsize
\begin{split}
& {\mathcal{L}_{DA}} = \frac{1}{C}\sum\limits_{c = 1}^C {{{\left( {{\mathcal{D}_{avg}}\left( {{\mathcal{S}_c}} \right) - {\alpha _c}} \right)}^2}}  - \frac{1}{C}\sum\limits_{c = 1}^C {{\alpha _c}} \\
& + \frac{1}{{{C^2}}}\sum\limits_{1 \le {c_i},{c_j} \le C} {{{\left( {{{\left( {\mathcal{D}_{avg}^{\left( 0 \right)}\left( {{\mathcal{S}_{{c_j}}}} \right)} \right)}^\eta } \cdot {\alpha _{{c_i}}} - {{\left( {\mathcal{D}_{avg}^{\left( 0 \right)}\left( {{\mathcal{S}_{{c_i}}}} \right)} \right)}^\eta } \cdot {\alpha _{{c_j}}}} \right)}^2}}.
\end{split}
\end{equation}
By minimizing the converted soft penalty term in Eq.(\ref{Eq:Eq8}), the inherent inter-class density correlations can be preserved in the learnt embedding space.

\subsection{Training Procedure}
Without loss of generality, we adopt the widely used contrastive embedding as the basic DML model and present how to additionally incorporate the density regularizer into it. It is also worth noting that our density regularizer is pluggable to any neural networks for deep metric learning and could be trained in an end-to-end fashion. In particular, the overall objective function of DML-DA integrates the contrastive loss in Eq.(\ref{Eq:Eq2}) and the proposed density regularizer in Eq.(\ref{Eq:Eq8}). Hence, we obtain the following optimization problem as
\begin{equation}\label{Eq:Eq9}
\mathop {\min }\limits_{\theta ,\left\{ {{\alpha _c}} \right\}_{c = 1}^C} {\mathcal{L}_{pair}} + \lambda  \cdot {\mathcal{L}_{DA}},
\end{equation}
where $\lambda$ is the tradeoff parameter. With this overall loss objective, the crucial goal of its optimization is to learn the embedding function $f\left( {x_i;\theta } \right)$ with its parameters $\theta$ and the target density of each class $\left\{ {{\alpha _c}} \right\}_{c = 1}^C$.

Inspired by the success of CNNs in recent DML models, we employ a deep architecture, i.e., GoogleNet \cite{Szegedy14}, followed by an additional fully-connected layer (an embedding layer) to learn the embedding representations for images. In the training stage, to solve the optimization according to overall loss objective in Eq.(\ref{Eq:Eq9}), we design a DML loss layer with density regularizer on the top of the embedding layer. The loss layer only contains parameters of target density. During learning, it evaluates the model's violation of both the basic DML supervision over pairs and density regularizer, and back-propagates the gradients with respect to target density of each class and input embedding representations to update the parameters of loss layer and lower layers, respectively. The training process of our DML-DA is given in Algorithm \ref{ag:ag01}.

\begin{table*}[!tb]
\centering
\caption{Performance comparisons with the state-of-the-art methods in terms of NMI and Recall@K (\%) on Cars196, CUB-200-2011 and Stanford Online Products dataset. The performances of Triplet (Tri), N-pair (NP) and Contrastive (Con) are reported based on our implementations and we utilize the models shared by the authors for HDC evaluation. The best performances are in bold and we also underline the performances of the best competitors. For the methods of Lifted Struct (LS) and Clustering (Clu), we directly extract results reported in \cite{song2016learnable}.}
\label{table:AP}
\begin{tabular}{@{~~}c@{~~~}|@{~~~}c@{~~~~}c@{~~~~}c@{~~~~}c@{~~~~}c@{~~~~}c@{~~~}|@{~~~}c@{~~~}c@{~~~}c@{~~}}
\hline
\multicolumn{10}{c}{\textbf{Cars196}}                                                                                              \\ \hline
\textbf{Method} & Tri \cite{Weinberger:NIPS06} & LS \cite{Song:CVPR16} & NP \cite{Sohn:NIPS16} & Clu \cite{song2016learnable} & Con \cite{bell15productnet} & HDC \cite{yuan2017hard}  & DML-DA$_{tri}$ & DML-DA$_{np}$ & DML-DA$_{con}$ \\ \hline
NMI    & 47.23   & 56.88        & 57.29  & 59.04      & 59.09       & \underline{62.17}     & 56.59      & 62.07     & \textbf{65.17}          \\ \hline
R@1    & 42.54   & 52.98        & 56.52  & 58.11      & 67.95       & \underline{71.42}     & 62.51      & 71.34     & \textbf{77.62}          \\
R@2    & 53.94   & 65.70        & 68.42  & 70.64      & 78.05       & \underline{81.85}     & 73.58      & 81.29     & \textbf{86.25}          \\
R@4    & 65.74   & 76.01        & 78.01  & 80.27      & 85.78       & \underline{88.54}     & 82.24      & 87.92     & \textbf{91.71}          \\
R@8    & 75.06   & 84.27        & 85.70  & 87.81      & 91.60       & \underline{93.40}     & 88.56      & 92.74     & \textbf{95.35}          \\
R@16   & 82.40   & -            & 91.19  & -          & 95.34       & \underline{96.59}     & 93.17      & 95.89     & \textbf{97.54}          \\
R@32   & 88.70   & -            & 94.81  & -          & 97.58       & \underline{98.16}     & 95.89      & 97.70     & \textbf{98.89}          \\
R@64   & 93.17   & -            & 97.38  & -          & 98.78       & \underline{99.21}     & 97.86	    & 98.82     & \textbf{99.37}          \\
R@128  & 96.42   & -            & 98.83  & -          & 99.51       & \underline{99.67}     & 98.98      & 99.53     & \textbf{99.73}          \\ \hline

\multicolumn{10}{c}{}\\
\hline
\multicolumn{10}{c}{\textbf{CUB-200-2011}}                                                                                                 \\ \hline
\textbf{Method} & Tri \cite{Weinberger:NIPS06} & LS \cite{Song:CVPR16} & NP \cite{Sohn:NIPS16} & Clu \cite{song2016learnable} & Con \cite{bell15productnet} & HDC \cite{yuan2017hard}  & DML-DA$_{tri}$ & DML-DA$_{np}$ & DML-DA$_{con}$ \\ \hline
NMI    & 50.99   & 56.50        & 57.41  & 59.23      & 60.07       & \underline{60.78}     & 55.53      & 59.67     & \textbf{62.32}   \\ \hline
R@1    & 39.57   & 43.57        & 47.30  & 48.18      & 52.01       & \underline{52.50}     & 45.90      & 51.45     & \textbf{55.64}   \\
R@2    & 51.74   & 56.55        & 59.57  & 61.44      & 65.16       & \underline{65.25}     & 57.97      & 63.01     & \textbf{66.96}   \\
R@4    & 63.35   & 68.59        & 70.75  & 71.83      & 75.71       & \underline{76.01}     & 69.53      & 74.38     & \textbf{77.92}   \\
R@8    & 74.14   & 79.63        & 80.98  & 81.92      & 84.25       & \underline{85.03}     & 80.23      & 83.78     & \textbf{86.23}   \\
R@16   & 82.98	& -	       & 88.28	& -	     & 90.82	   & \underline{91.10}     & 88.15	    & 90.58	& \textbf{92.10}   \\
R@32   & 89.53	& -	       & 93.50	& -	     & 95.17	   & \underline{95.34}     & 94.01	    & 95.16	& \textbf{95.95}   \\
R@64   & 94.80	& -	       & 96.79	& -	     & \underline{97.70} & 97.67           & 97.00	    & 97.64	& \textbf{98.11}   \\
R@128  & 97.65	& -	       & 98.43	& -	     & 98.99	   & \underline{99.09}     & 98.63	    & 98.89	& \textbf{99.21}   \\ \hline

\multicolumn{10}{c}{}\\
\hline
\multicolumn{10}{c}{\textbf{Stanford Online Products}}                                                                                     \\ \hline
\textbf{Method} & Tri \cite{Weinberger:NIPS06} & LS \cite{Song:CVPR16} & NP \cite{Sohn:NIPS16} & Clu \cite{song2016learnable} & Con \cite{bell15productnet} & HDC \cite{yuan2017hard}  & DML-DA$_{tri}$ & DML-DA$_{np}$ & DML-DA$_{con}$ \\ \hline
NMI    & 86.20   & 88.65        & 88.77  & \underline{89.48}      & 88.57       & 88.75    & 87.25      & 88.93     & \textbf{89.50}   \\  \hline
R@1    & 59.49   & 62.46        & 65.89  & 67.02      & 68.20       & \underline{69.17}    & 61.16      & 67.49     & \textbf{70.56}   \\
R@10   & 76.23   & 80.81        & 81.94  & \underline{83.65}      & 82.20       & 82.77    & 78.99      & 82.20     & \textbf{84.09}   \\
R@100  & 87.95   & 91.93        & 91.83  & \underline{93.23}      & 90.87       & 91.27    & 90.54      & 91.94     & \textbf{94.09}   \\
R@1000 & 95.70	& -	       & 97.30	& -	     & 96.61	   & \underline{97.59}    & 96.96	   & 97.69	   & \textbf{97.72}   \\ \hline
\end{tabular}
\vspace{-0.00in}
\end{table*}

\section{Experiments}\label{sec:EX}
We evaluate our DML-DA models by conducting two object recognition tasks (clustering and $k$-nearest neighbour retrieval) on three image datasets, i.e., Cars196 \cite{Krause:ICCV13}, CUB-200-2011 \cite{wah2011caltech} and Stanford Online Products \cite{Song:CVPR16}. The first two are the popular fine-grained object recognition benchmarks and the latter one is a recently released object recognition dataset of online product images.

\textbf{Cars196} contains 16,185 images belonging to 196 classes of cars. In our experiments, we follow the settings in \cite{Song:CVPR16}, taking the first 98 classes (8,054 images) for training and the rest 98 classes (8,131 images) for testing.

\textbf{CUB-200-2011} includes 11,788 images of 200 classes corresponding to different birds species. Following \cite{Song:CVPR16}, we utilize the first 100 classes (5,864 images) for training and the remaining 100 classes (5,924 images) for testing.

\textbf{Stanford Online Products} is a recent collection of online product images from eBay.com. It is composed of 120,053 images belonging to 22,634 classes. In our experiments, we utilize the standard split in \cite{Song:CVPR16}: 11,318 classes (59,551 images) are used for training and 11,316 classes (60,502 images) are exploited for testing.

\subsection{Implementation Details}
For the network architecture, we utilize GoogleNet \cite{Szegedy14} pre-trained on Imagenet ILSVRC12 dataset \cite{ILSVRC15} plus a fully connected layer (an embedding layer), which is initialized with random weights. For density regularizer, its parameters (i.e., the target density of each class) are all initially set to 0.5. The control factor $\eta$ in Eq.(\ref{Eq:Eq7}) is set as 0.5 and the tradeoff parameter $\lambda$ in Eq.(\ref{Eq:Eq9}) is fixed to 10. All the margin parameters (e.g., $m_p$ and $m_t$) are set to 1. We fix the embedding size $d$ as 128 throughout the experiments. We mainly implement DML models based on Caffe \cite{Jia:MM14}, which is one of widely adopted deep learning frameworks. Specifically, the network weights are trained by ADAM \cite{kingma2014adam} with 0.9/0.999 momentum. The learning rate is initially set as $5 \times 10^{-4}$, $10^{-5}$ and $3 \times 10^{-5}$ on Cars196, CUB-200-2011 and Stanford Online Products, respectively. The mini-batch size is set as 100 and the maximum training iteration is set as 30,000 for all the experiments. In the experiments on Cars196 and CUB-200-2011, to compute the density of each class with sufficient images in a mini-batch, we first randomly sample 10 classes from all training classes and then randomly select 10 images for each sampled class, leading to the mini-batch with 100 training images. In the experiments on Stanford Online Products dataset, since each training class contains only 5 images on average, we construct each mini-batch by accumulating all the images for randomly sampled classes until the maximum size of mini-batch is achieved.

\begin{figure*}[!tb]
   \centering
   \subfigure[]{
     \label{fig:fig3:a}
     \includegraphics[width=0.32\textwidth]{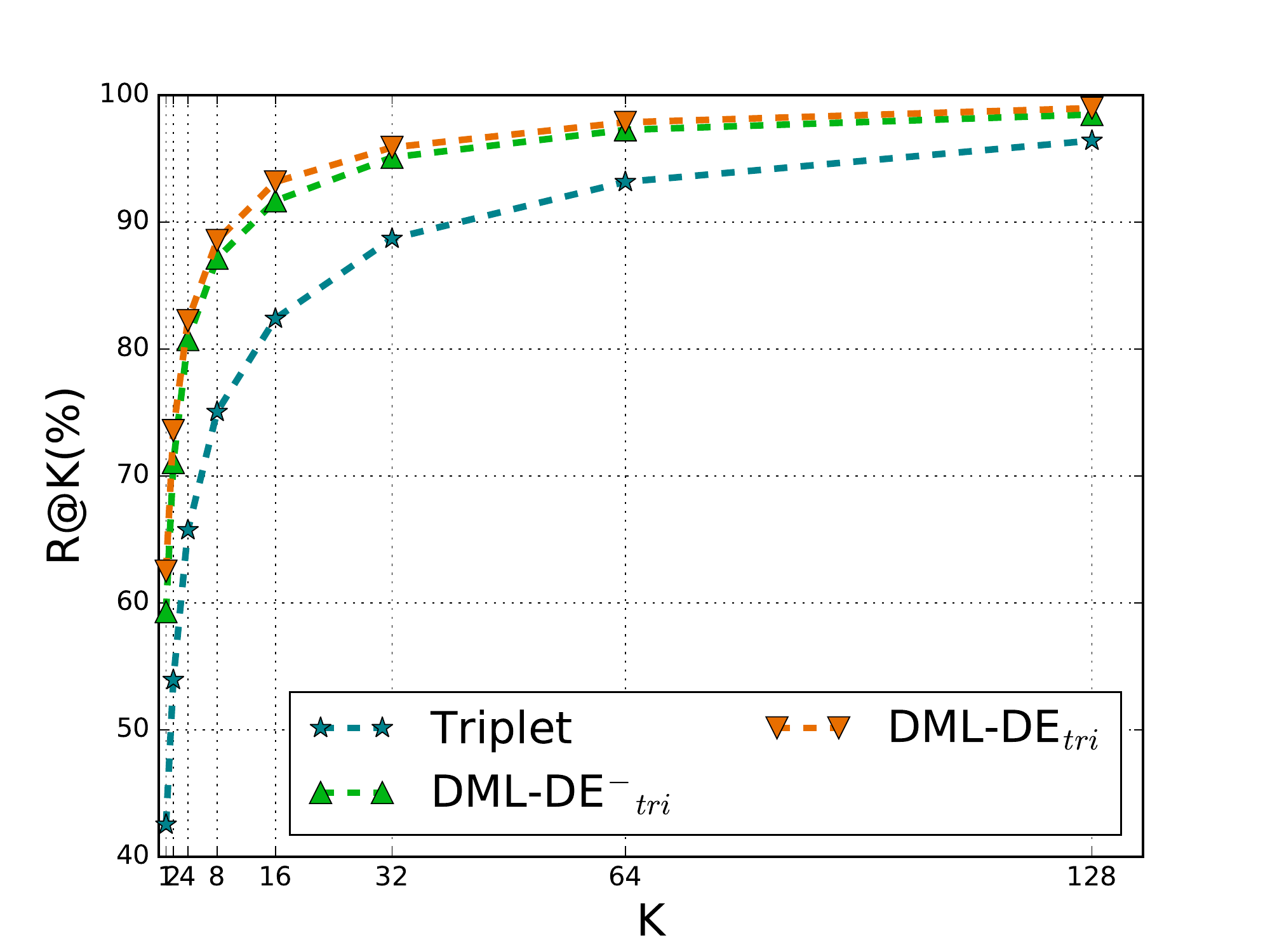}}
   \subfigure[]{
     \label{fig:fig3:b}
     \includegraphics[width=0.32\textwidth]{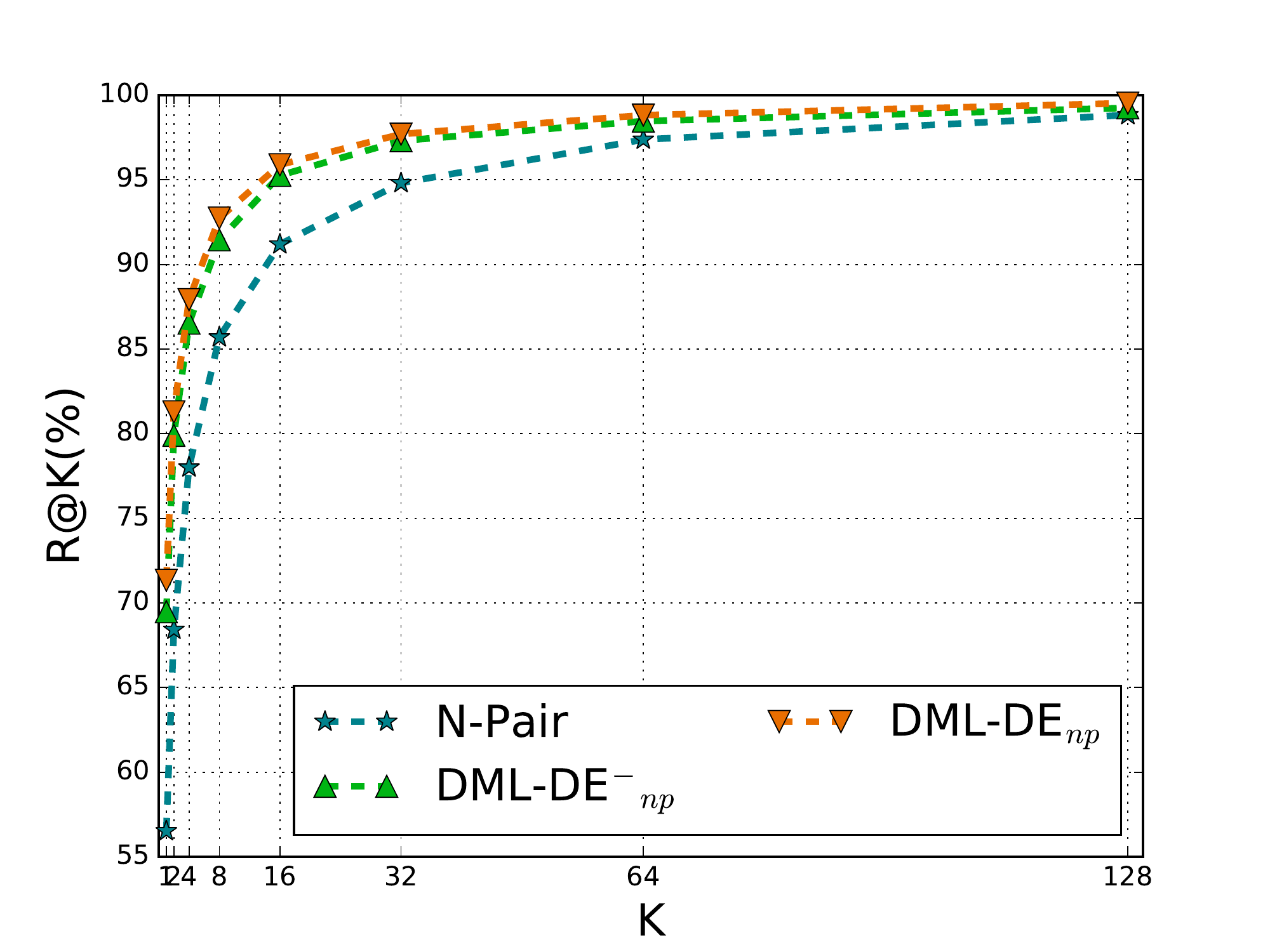}}
   \subfigure[]{
     \label{fig:fig3:c}
     \includegraphics[width=0.32\textwidth]{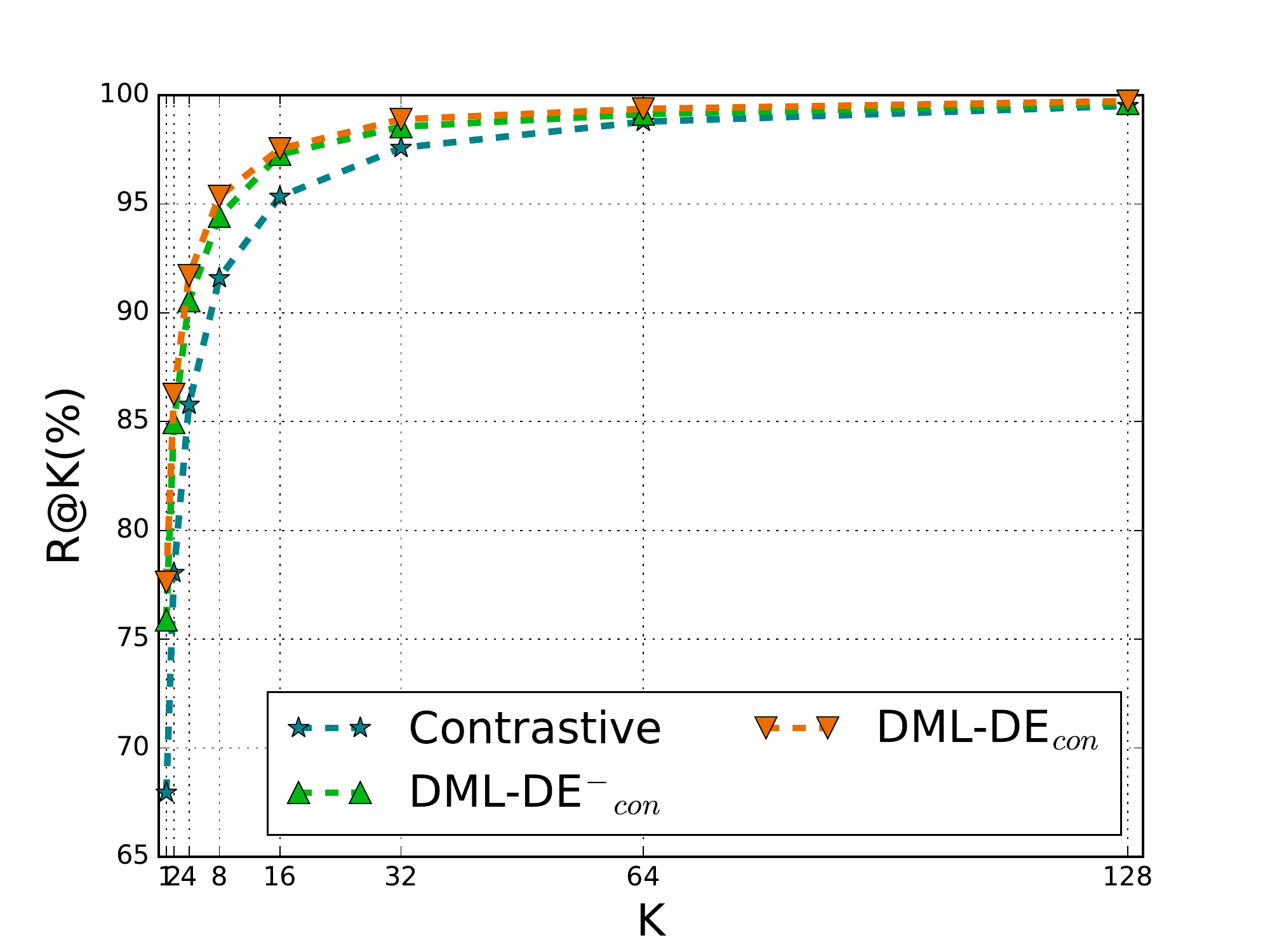}}
     \vspace{-0.00in}
   \caption{Recall@K performance comparison of DML-DA framework w or w/o inter-class density correlations preservation on Cars196 dataset, when exploiting (a) triplet loss, (b) N-pair loss and (c) contrastive loss.}
   \label{fig:fig3}
   \vspace{-0.1in}
\end{figure*}

\begin{figure*}[!tb]
   \centering
   \subfigure[]{
     \label{fig:fig5:a}
     \includegraphics[width=0.32\textwidth]{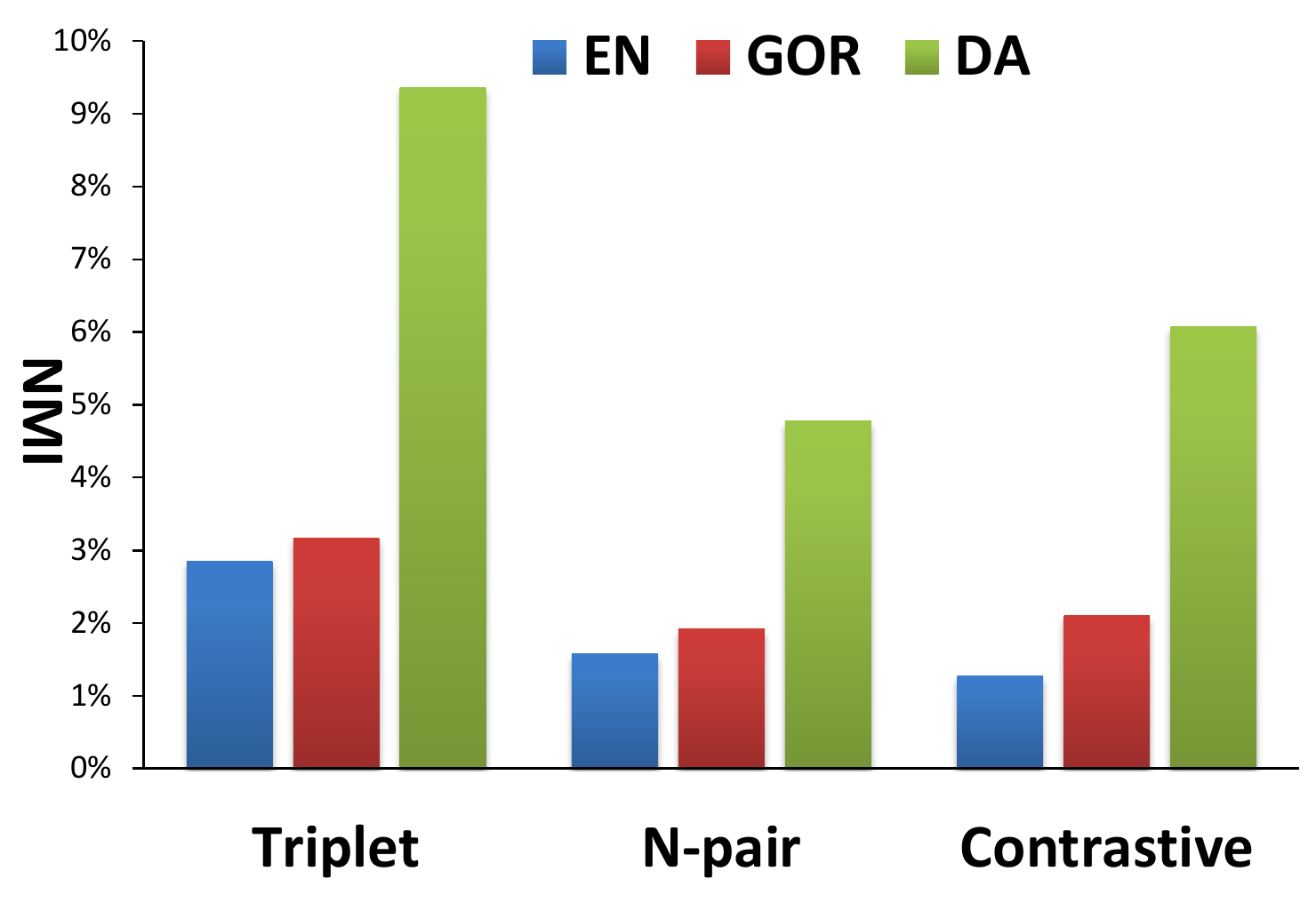}}
   \subfigure[]{
     \label{fig:fig5:b}
     \includegraphics[width=0.31\textwidth]{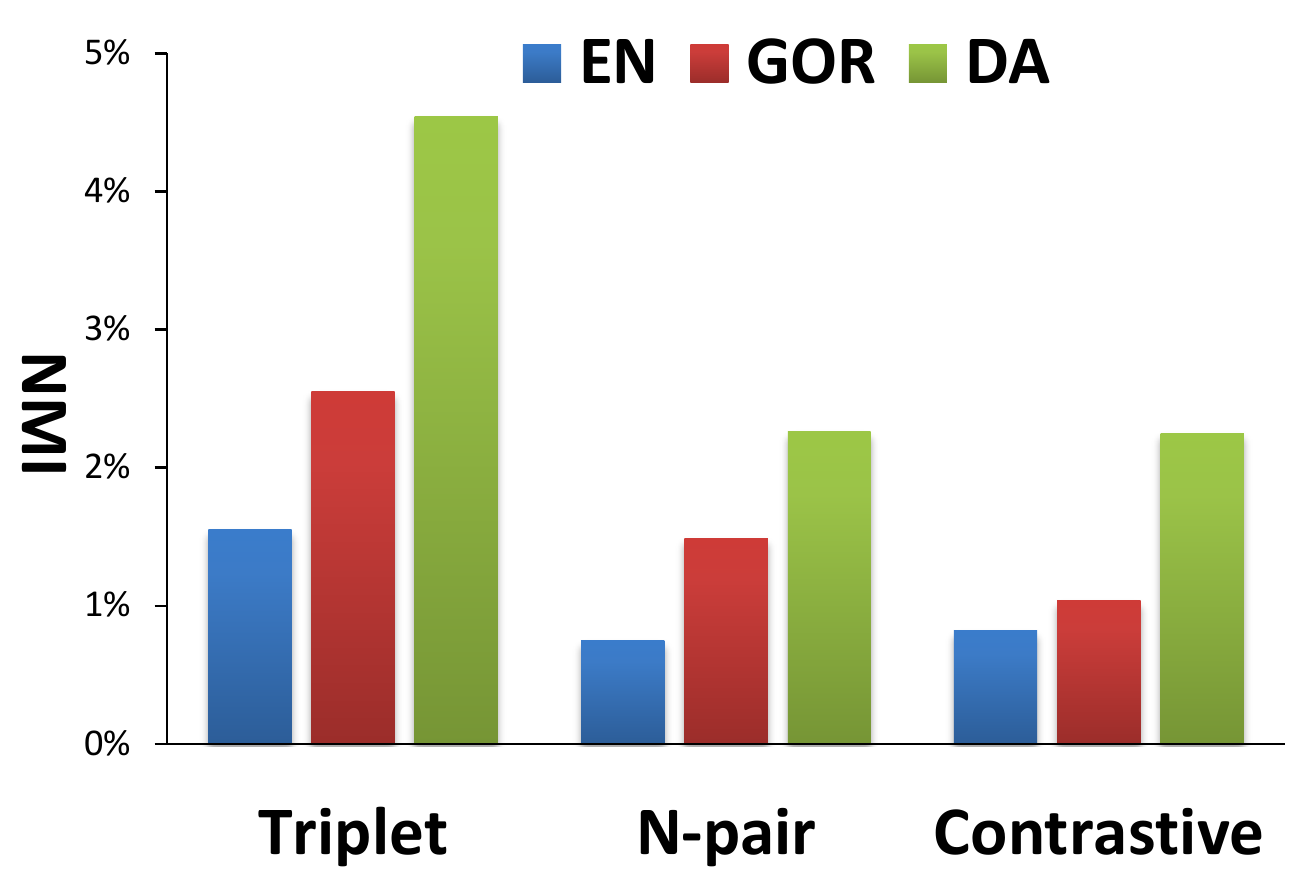}}
   \subfigure[]{
     \label{fig:fig5:c}
     \includegraphics[width=0.32\textwidth]{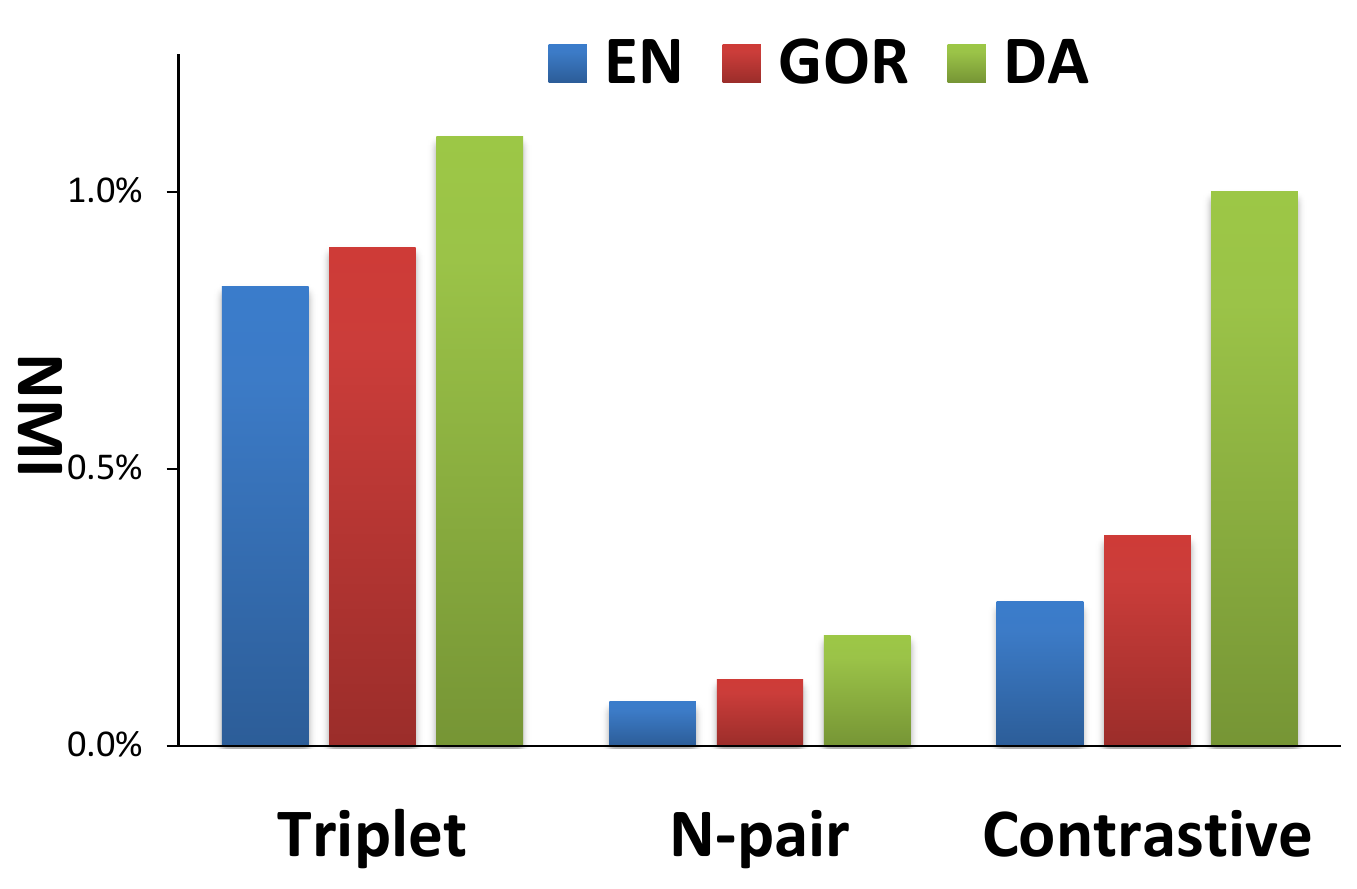}}
     \vspace{-0.000in}
   \caption{NMI performance gains when plugging entropy (EN) regularizer, global orthogonal (GOR) regularizer and our density adaptivity (DA) regularizer into DML architecture with triplet loss, N-pair loss and contrastive loss, on (a) Car196, (b) CUB-200-2011 and (c) Stanford Online Products.}
   \label{fig:fig5}
   \vspace{-0.00in}
\end{figure*}

\subsection{Evaluation Metrics and Compared Methods}
\textbf{Evaluation Metrics.} For the clustering task, we adopt the Normalised Mutual Information (NMI) \cite{manning2008introduction} metric, which is defined as the ratio of mutual information and the average entropy of clusters and labels. For the $k$-nearest neighbour retrieval task, Recall@K (R@K) is utilized for quantitative evaluation. Given a test image query, its Recall@K score is measured as 1 if an image of the same class is retrieved among the $k$-nearest neighbours and 0 otherwise. The final metric score is the average of Recall@K for all image queries in the testing set. All the metrics are computed by using the codes\footnote{\small\url {https://github.com/rksltnl/Deep-Metric-Learning-CVPR16/tree/master/code/evaluation}} released in \cite{Song:CVPR16}.

\textbf{Compared Methods.} To empirically verify the merit of our proposed DML-DA models, we compared the following state-of-the-art methods:

(1) \textbf{Triplet} \cite{Weinberger:NIPS06} adopts triplet loss to optimize the deep architecture. (2) \textbf{Lifted Struct} \cite{Song:CVPR16} devises a structured prediction objective on the lifted dense pairwise distance matrix within the batch. (3) \textbf{N-Pair} \cite{Sohn:NIPS16} trains DML with N-pair loss. (4) \textbf{Clustering} \cite{song2016learnable} is a structured prediction based DML model which can be optimized with clustering quality metric. (5) \textbf{Contrastive} \cite{bell15productnet} uses contrastive loss for DML training. (6) \textbf{HDC} \cite{yuan2017hard} trains the embedding neural network in a cascaded manner by handling samples of different hard level with models of different complexities. (7) \textbf{DML-DA} is the proposal in this paper. \textbf{DML-DA$_{tri}$}, \textbf{DML-DA$_{np}$} and \textbf{DML-DA$_{con}$} denotes that the basic DML model in our DML-DA is equipped with triplet loss, N-pair loss and contrastive loss, respectively. Moreover, a slightly different settings of the three runs are named as \textbf{DML-DA$^{-}_{tri}$}, \textbf{DML-DA$^{-}_{np}$} and \textbf{DML-DA$^{-}_{con}$}, which are all trained without inter-class density correlations preservation constraint.

\subsection{Performance Comparison}
Table \ref{table:AP} shows the NMI and k-nearest neighbors performance with Recall@K metric of different approaches on Cars196, CUB-200-2011 and Stanford Online Products dataset, respectively. It is worth noting that the dimension of the embedding space in Triplet, N-pair, Contrastive, HDC and our three DML-DA runs is 128, and in Lifted Struct and Clustering, the performances are given by choosing 64 as the embedding dimension. In view that the embedding size is not sensitive towards performance during training and testing phase as studied in \cite{Song:CVPR16}, we compare directly with results.

Overall, the results across all evaluation metrics (NMI and Recall at different depths) and three datasets consistently indicate that our proposed DML-DA$_{con}$ exhibits better performance against all the state-of-the-art techniques. In particular, the NMI and Recall@1 performance of DML-DA$_{con}$ can achieve 65.17\% and 77.62\%, making the absolute improvement over the best competitor HDC by 3.0\% and 6.2\% on Cars196, respectively. DML-DA$_{tri}$, DML-DA$_{np}$ and DML-DA$_{con}$ by integrating density adaptivity makes the absolute improvement over Triplet, N-pair and Contrastive by 19.97\%, 14.82\% and 9.67\% in Recall@1 on Cars196, respectively. The performance trends on the other two datasets are similar with that of Cars196. The results indicate the advantage of exploring density adaptivity in DML training to enhance model generalization. Triplet which only compares an example with one negative example while ignoring negative examples from the rest of the classes performs the worst among all the methods. Lifted Struct, N-pair and Clustering distinguishing an example from all the negative classes lead to a large performance boost against Triplet.

\begin{figure*}[!tb]
   \centering
   \subfigure[Triplet]{
     \label{fig:fig4:a}
     \includegraphics[width=0.29\textwidth]{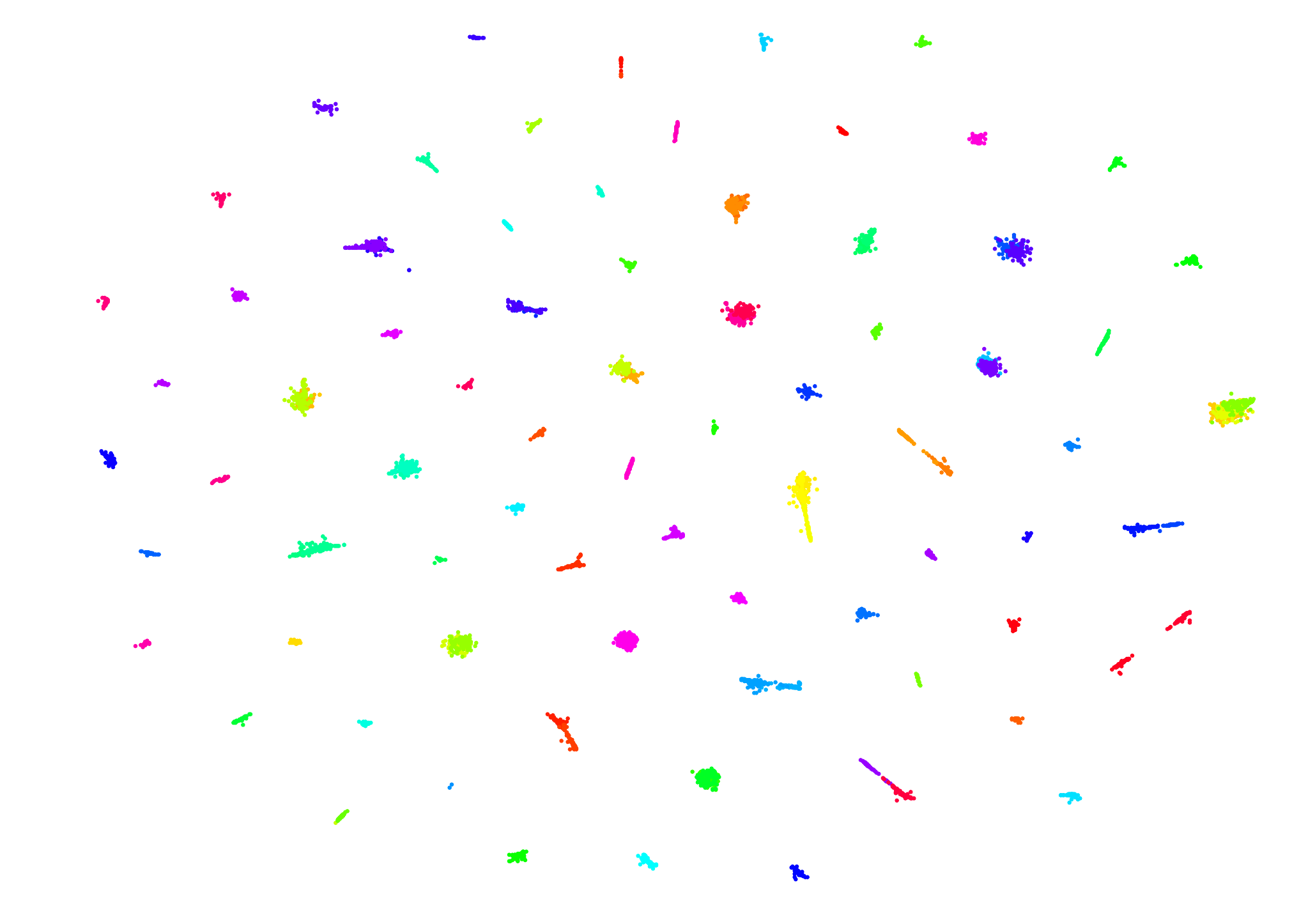}}
     \hspace{0mm}
   \subfigure[N-pair]{
     \label{fig:fig4:b}
     \includegraphics[width=0.3\textwidth]{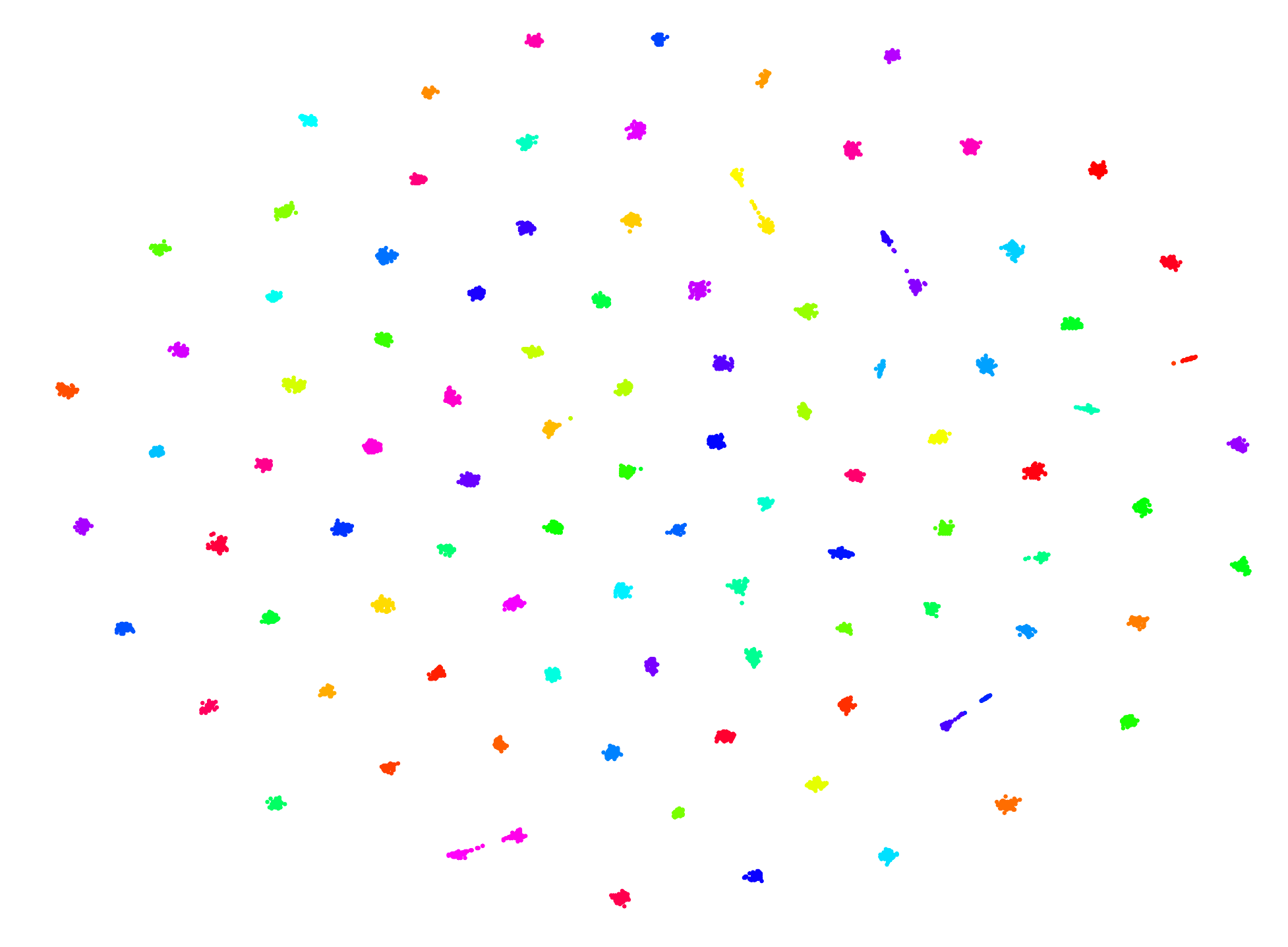}}
     \hspace{0mm}
   \subfigure[Contrastive]{
     \label{fig:fig4:c}
     \includegraphics[width=0.3\textwidth]{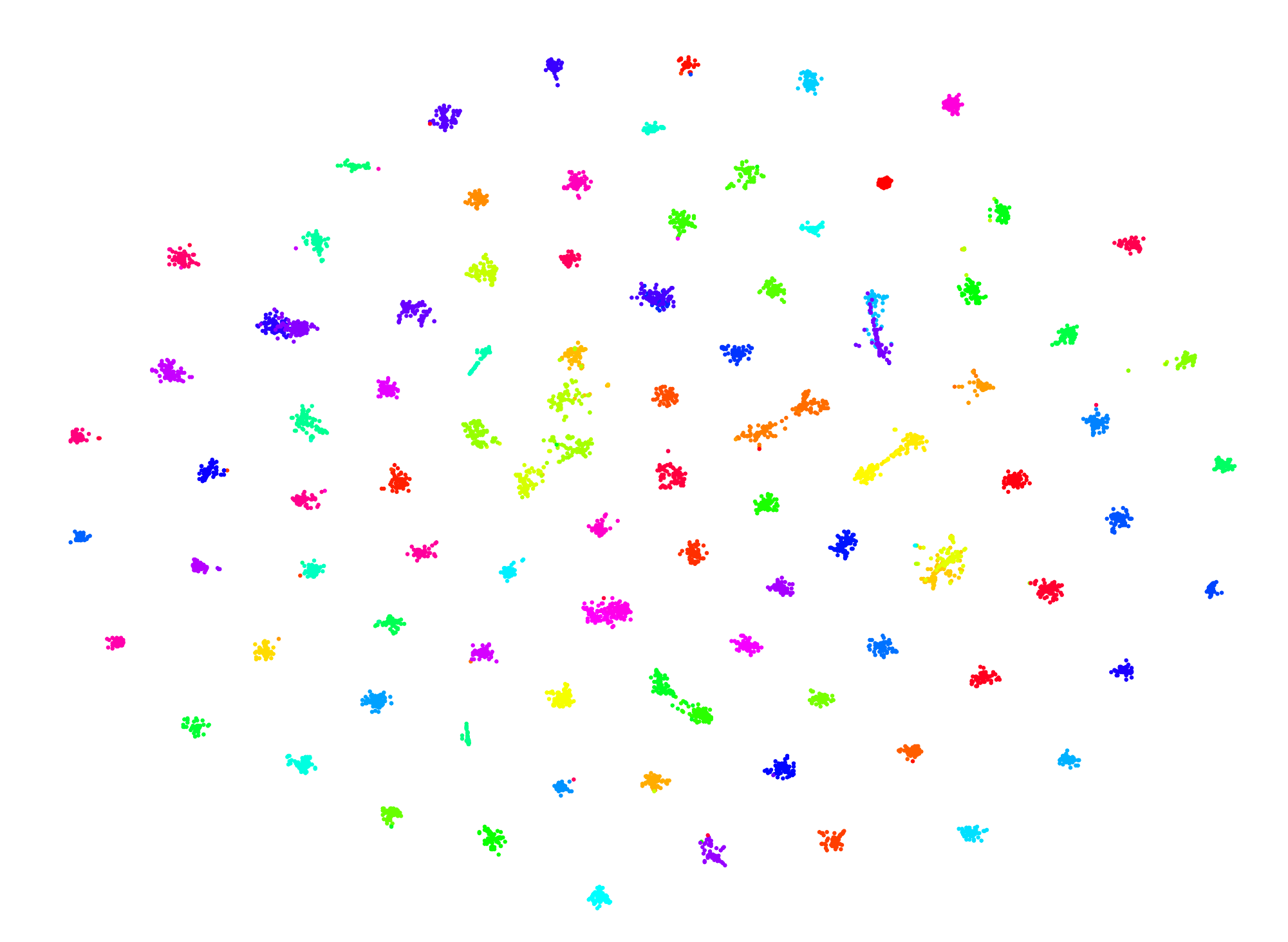}}

     \vspace{-0.0in}

   \subfigure[DML-DA$_{tri}$]{
     \label{fig:fig4:d}
     \includegraphics[width=0.3\textwidth]{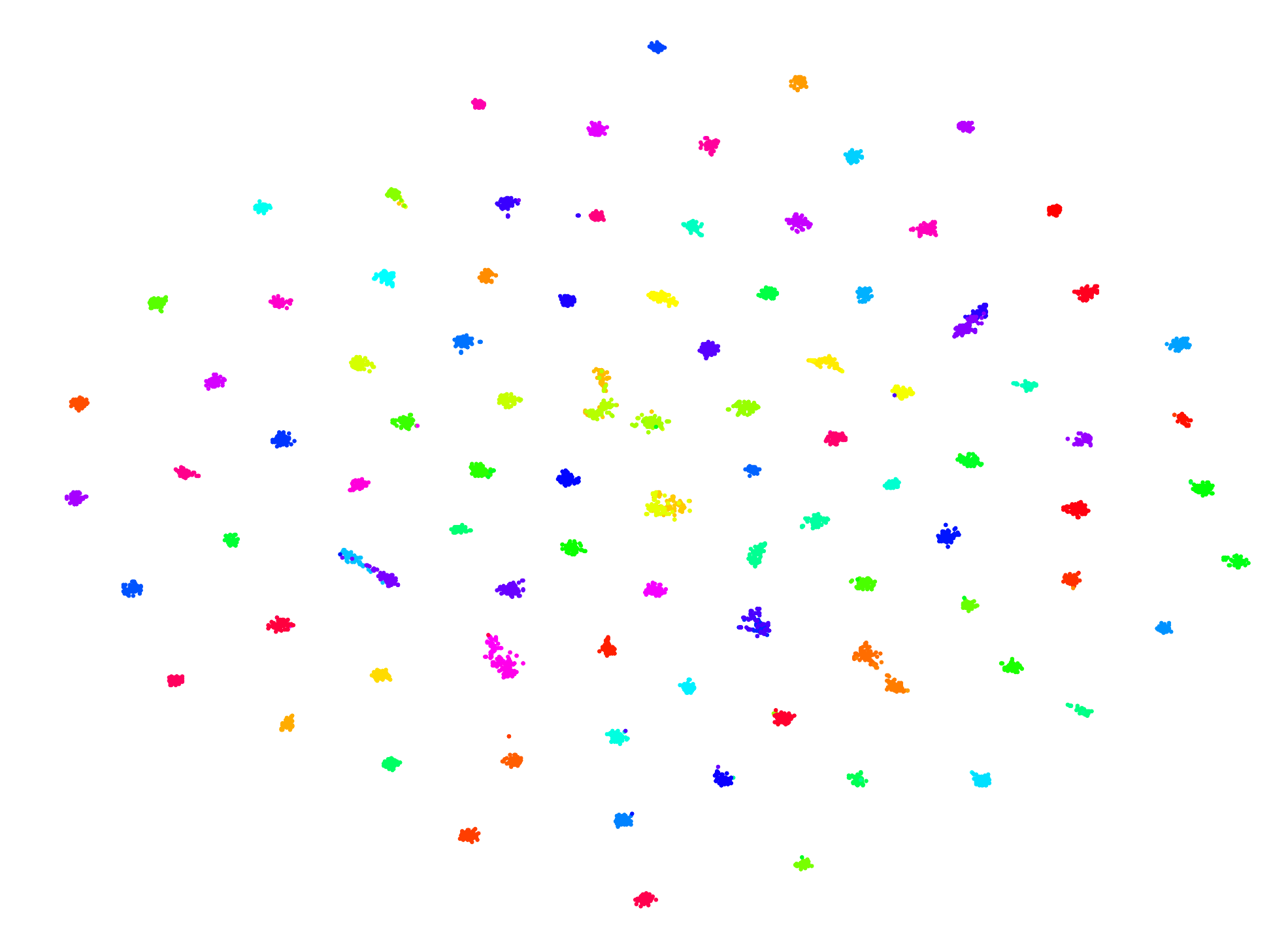}}
     \hspace{0mm}
   \subfigure[DML-DA$_{np}$]{
     \label{fig:fig4:e}
     \includegraphics[width=0.3\textwidth]{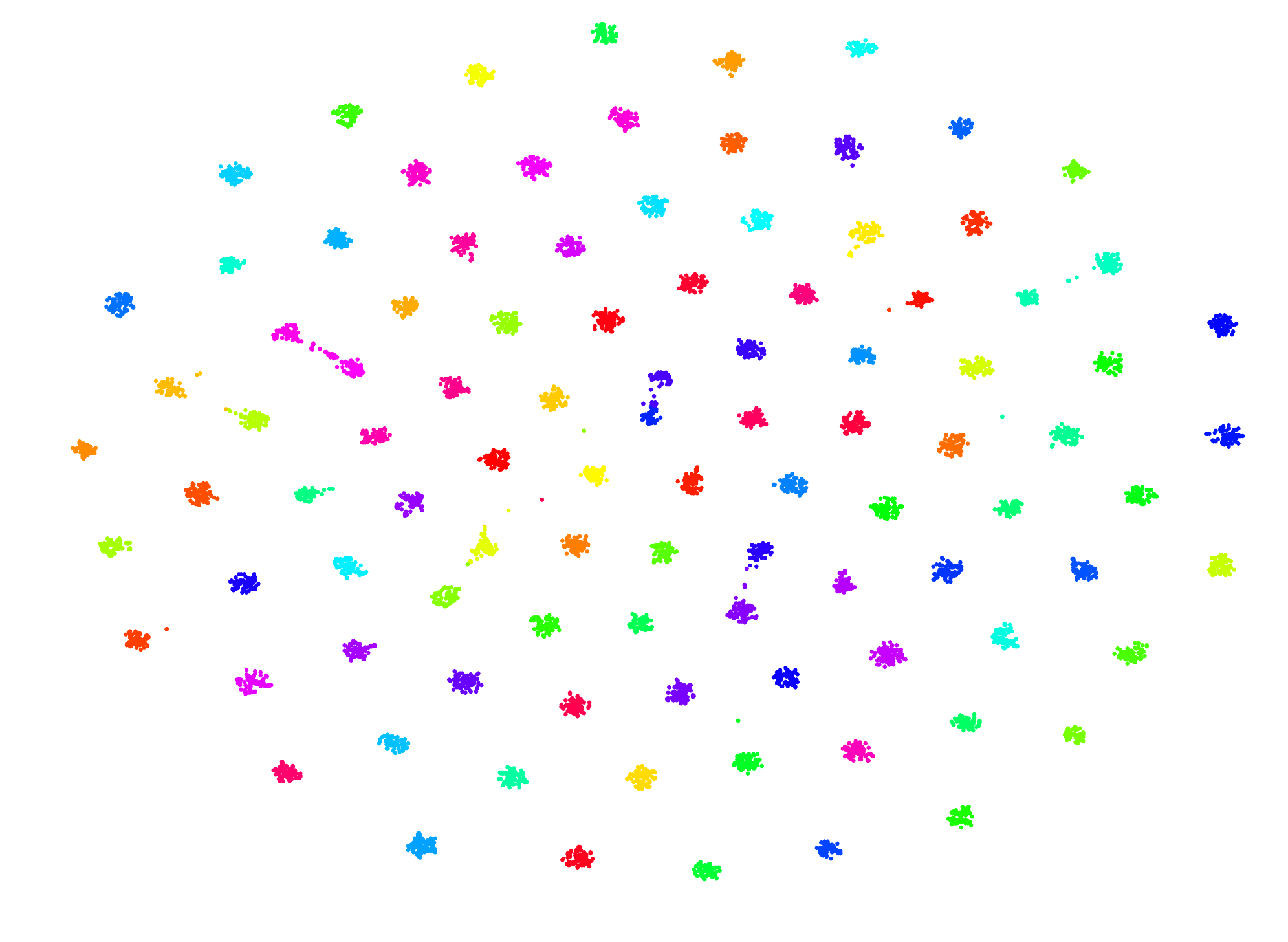}}
     \hspace{0mm}
  \subfigure[DML-DA$_{con}$]{
     \label{fig:fig4:f}
     \includegraphics[width=0.3\textwidth]{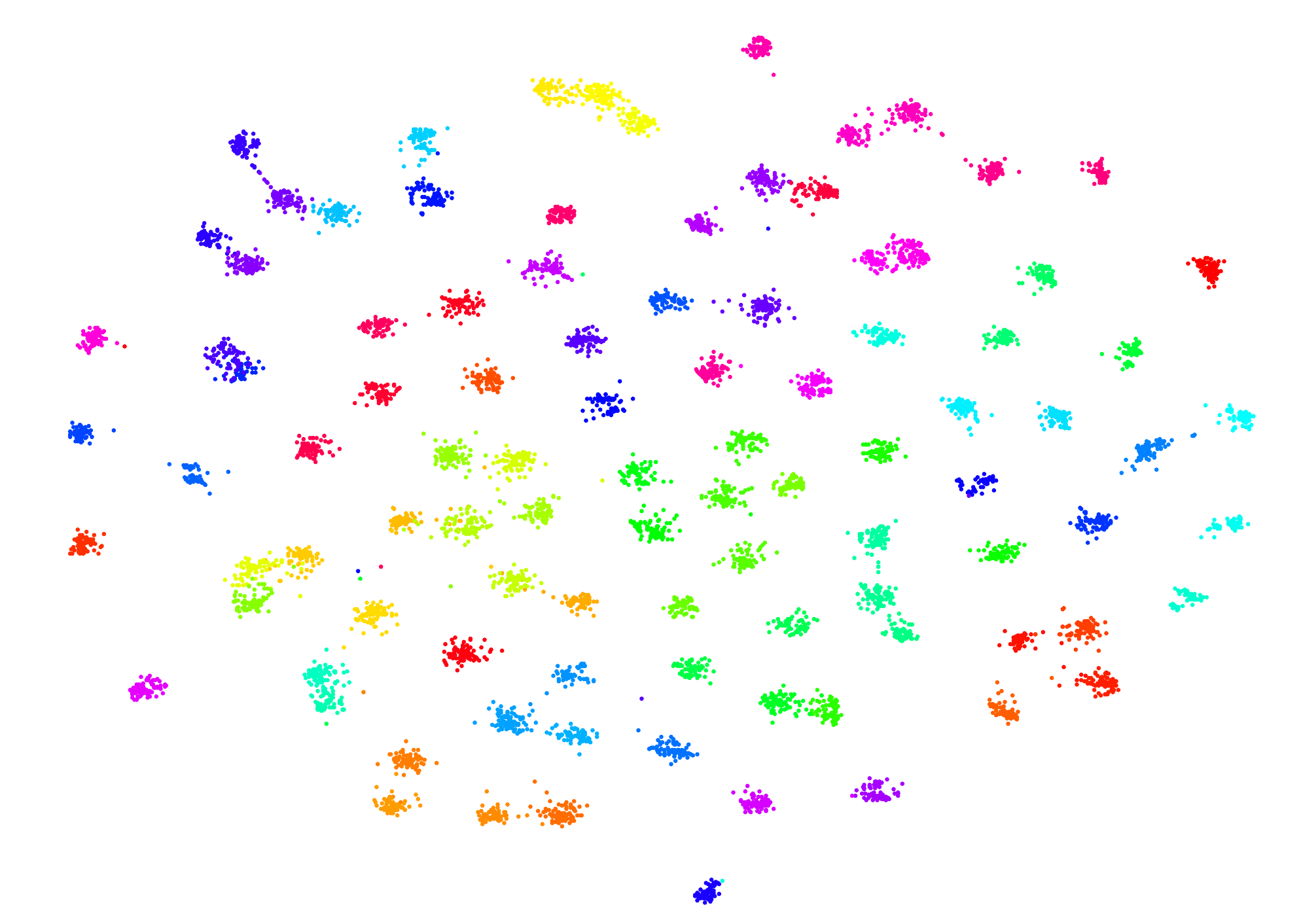}}
   \caption{Image representation embedding visualizations of the training 98 classes in Cars196 by using t-SNE \cite{maaten:JMLR08}. Each image is visualized as one point and colors denote different classes. The embedding space is learnt by Triplet, N-pair, Contrastive, our DML-DA$_{tri}$, DML-DA$_{np}$ and DML-DA$_{con}$, respectively.}
   \label{fig:fig4}
   \vspace{-0.00in}
\end{figure*}

\begin{figure*}[!tb]
   \centering {\includegraphics[width=0.85\textwidth]{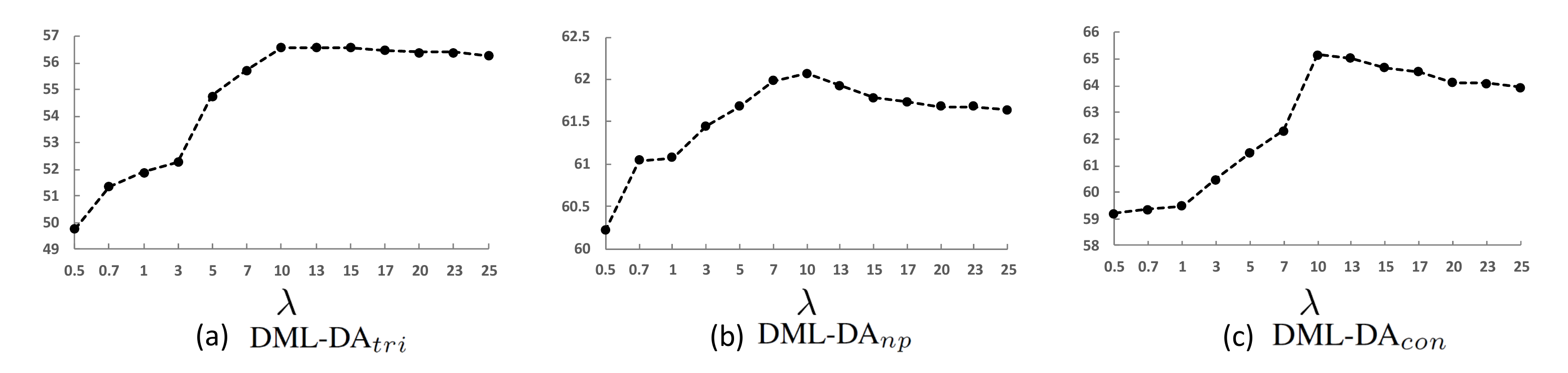}}
   \caption{The effect of the tradeoff parameter $\lambda$ in our (a) DML-DA$_{tri}$, (b) DML-DA$_{np}$ and (c) DML-DA$_{con}$ over NMI (\%) on Cars196 dataset.}
   \label{fig:figs_lamba}
   \vspace{-0.0in}
\end{figure*}

\begin{figure*}[!tb]
   \centering {\includegraphics[width=0.73\textwidth]{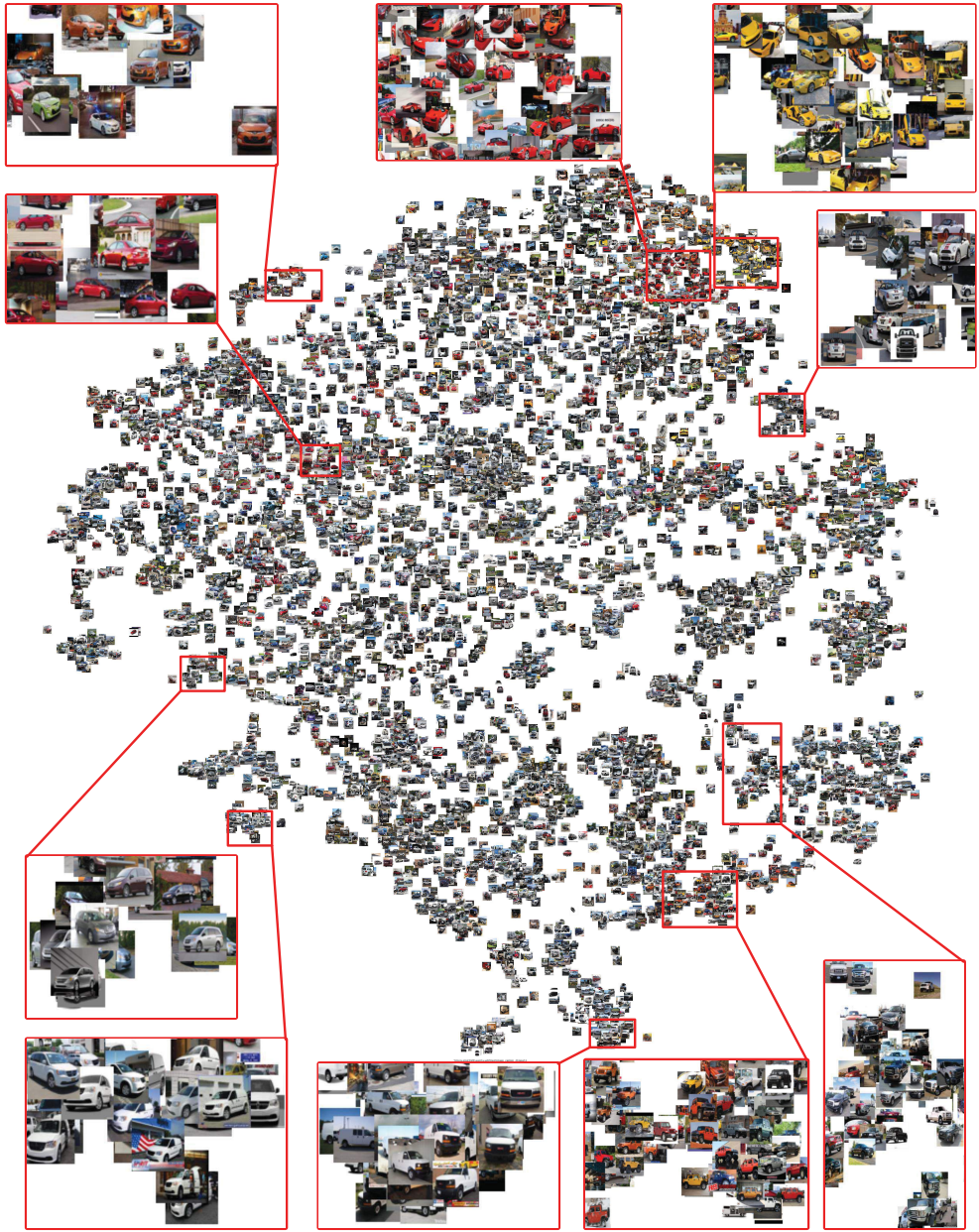}}
   \caption{Barnes-Hut t-SNE visualization \cite{van2014accelerating} of image embedding representations learnt by our DML-DA$_{con}$ on the test split of Cars196. Best viewed on a monitor when zoomed in. By integrating density adaptivity in DML training, our DML-DA$_{con}$ effectively balances the inter-class similarity and intra-class variation, which enhances model generalization. As such, the learnt embedding representation is more discriminative to cluster semantically similar cars despite of the significant variations in pose and body paint.}
   \label{fig:figs_example_car}
   \vspace{-0.0in}
\end{figure*}

Contrastive outperforms Lifted Struct, N-pair and Clustering on both Cars196 and CUB-200-2001. Though the four runs all involve the utilization of the relationship in both positive pairs and negative pairs, they are fundamentally different in devising the objective function that Lifted Struct, N-pair and Clustering tend to push positive pairs closer through negative pairs and encourage small intra-class variation, while Contrastive could flexibly balance inter-class distance and intra-class similarity by seeking a tradeoff of impact between positive pairs and negative pairs. As indicated by our results, advisably enlarging intra-class variation leads to better performance and makes Contrastive generalize well. This is also consistent with the motivation of our density adaptivity, which is to regularize the degree of data concentration of each class. With our density adaptivity, DML-DA$_{con}$ successfully boosts up the performance on the two datasets. In contrast, the NMI performance of Contrastive is inferior to that of Lifted Struct, N-pair and Clustering on Stanford Online Products. This is expected as the number of classes in Stanford Online Products is too large (more than 11K test classes) and thus Lifted Struct, N-pair and Clustering are benefited from the outcome of small intra-class clustering, making the chance of distinguishingly distributing such a large number of classes on the embedding space better. The improvement is also observed by DML-DA$_{con}$ in this extreme case. Furthermore, HDC by handling samples of different hard levels with sub-networks of different depths improves Contrastive, but the performances are still lower than our DML-DA$_{con}$.

\subsection{Effect of Inter-class Density Correlations Preservation}
Figure \ref{fig:fig3} compares the Recall@K performance of our DML-DA framework with or without inter-class density correlations preservation constraint on Cars196 dataset. The results across different depths (K) of Recall consistently indicate that additionally exploring inter-class density correlations preservation exhibits better performance when exploiting triplet loss, N-pair loss and contrastive loss in our DML-DA framework, respectively. Though the performance gain is gradually decreased when going deeper into the retrieval list, our DML-DA framework still leads to apparent improvement, even at Recall@128. In particular, DML-DA$_{tri}$ makes the absolute improvement over DML-DA$^{-}_{tri}$ and Triplet by 0.5\% and 2.56\% in terms of Recall@128, respectively.

\subsection{Effect of Different Regularizer}
Next, we compare our density regularizer with Entropy (EN) regularizer \cite{Niu:ICML12} and Global Orthogonal (GOR) regularizer \cite{Zhang:ICCV17} by plugging each of them into DML architecture with triplet loss, N-pair loss and contrastive loss, respectively. The entropy regularizer aims to maximize the entropy of the representation distribution in the embedding space and thus implicitly encourages large intra-class variation and small inter-class distance. The global orthogonal regularizer is to maximize the spread of embedding representations following the property that two non-matching representations are close to orthogonal with a high probability.

\begin{figure*}[!tb]
   \centering {\includegraphics[width=0.75\textwidth]{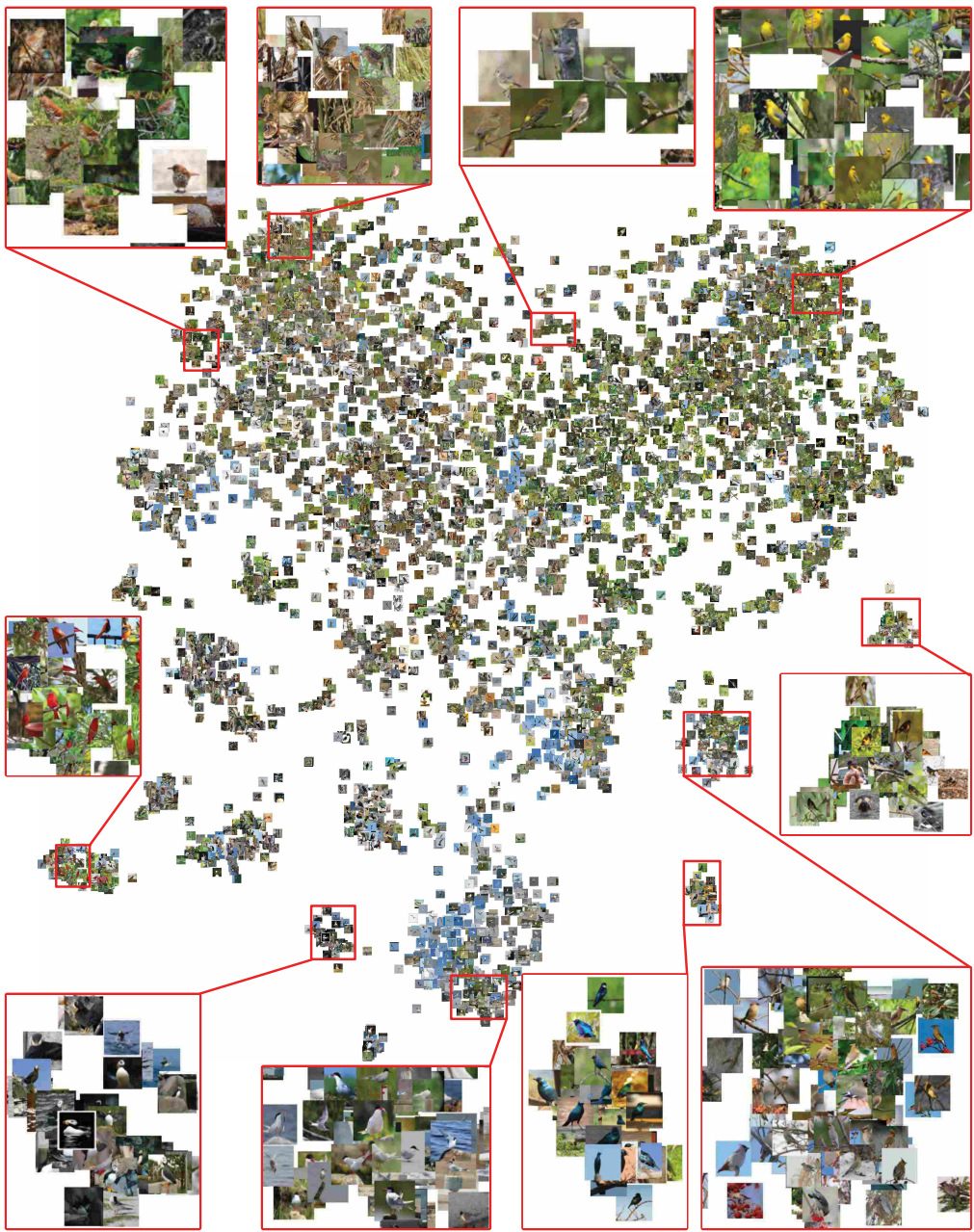}}
   \caption{Barnes-Hut t-SNE visualization \cite{van2014accelerating} of image embedding representations learnt by our DML-DA$_{con}$ on the test split of CUB-200-2011 dataset. Best viewed on a monitor when zoomed in. By integrating density adaptivity in DML training, our DML-DA$_{con}$ effectively balances the inter-class similarity and intra-class variation, which enhances model generalization. As such, the learnt embedding representation is more discriminative to cluster semantically similar birds despite of the significant variations in view point and background.}
   \label{fig:figs_example_bird}
   \vspace{-0.0in}
\end{figure*}

Figure \ref{fig:fig5} details the NMI performance gains when exploiting each of the three regularizers on Cars196, CUB-200-2011 and Stanford Online Products dataset, respectively. The results across DML architecture with three types of losses and three datasets consistently indicate that our DA regularizer leads to a larger performance boost against the other two regularizers. Compared to EN regularizer, our DA regularizer is more effective and robust, since we uniquely consider the balance between enlarging intra-class variation and penalizing distribution overlap across different classes in the optimization. GOR regularizer targeting for an uniform distribution of examples in the embedding space improves EN regularizer, but the performance is still lower than that of our DA regularizer. This somewhat reveals the weakness of GOR regularizer which performs a strong constraint of pushing two randomly examples from different categories close to orthogonal. In addition, the improvement trends on other evaluation metrics are similar with that of NMI.

\subsection{Effect of trade-off parameter $\lambda$}
To further clarify the effect of the tradeoff parameter $\lambda$ in Eq.(\ref{Eq:Eq9}), we illustrate the performance curves of DML-DA with three types of losses by varying $\lambda$ from 0.5 to 25 in Figure \ref{fig:figs_lamba}. As shown in the figure, our DML-DA architecture with three types of losses constantly indicate that the best NMI performance is attained when the tradeoff parameter $\lambda$ is set to 10. More importantly, the performance curve for each DML-DA model is relatively smooth as long as $\lambda$ is larger than 7, that practically eases the selection of $\lambda$.

\begin{figure*}[!tb]
   \centering {\includegraphics[width=0.75\textwidth]{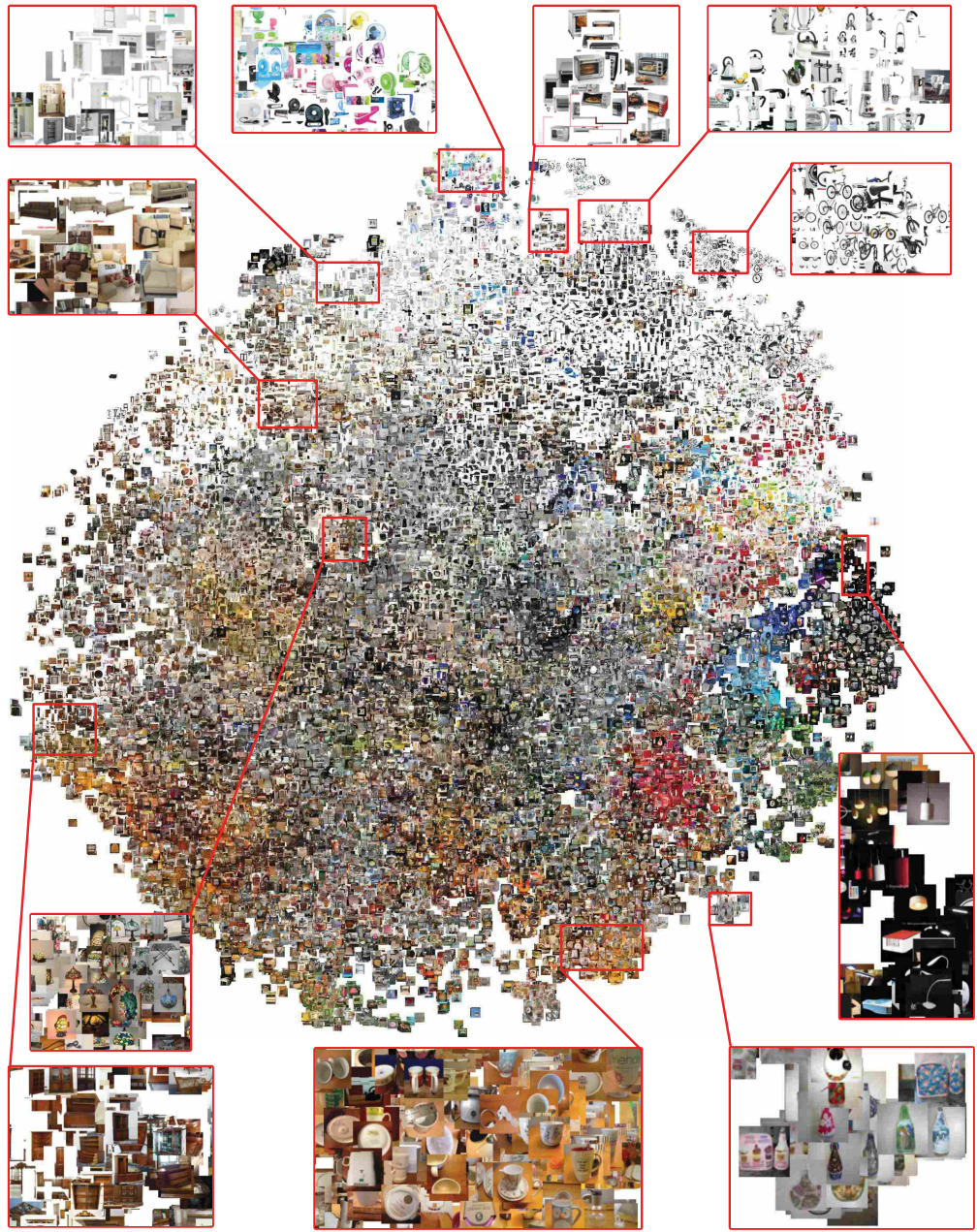}}
   \caption{Barnes-Hut t-SNE visualization \cite{van2014accelerating} of image embedding representations learnt by our DML-DA$_{con}$ on the test split of Stanford Online Products dataset. Best viewed on a monitor when zoomed in. By integrating density adaptivity in DML training, our DML-DA$_{con}$ effectively balances the inter-class similarity and intra-class variation, which enhances model generalization. As such, the learnt embedding representation is more discriminative to cluster semantically similar products despite of the significant variations in configuration and illumination.}
   \label{fig:figs_example_product}
   \vspace{-0.0in}
\end{figure*}

\subsection{Embedding Representations Visualization}
Figure \ref{fig:fig4:a}---\ref{fig:fig4:f} shows the t-SNE \cite{maaten:JMLR08} visualizations of image embedding representations learnt by Triplet, N-pair, Contrastive, our DML-DA$_{tri}$, DML-DA$_{np}$, and DML-DA$_{con}$, respectively. Specifically, we utilize all the training 98 classes in Cars196 dataset and the embedding representations of all the 8,054 images are then projected into 2-dimensional space using t-SNE. It is clear that the intra-class variation of the embedding representations learnt by DML-DA$_{tri}$ is larger than those of Triplet, while guaranteeing all the classes separable. Similarly, the increase of intra-class variation is also observed in t-SNE visualization when integrating density adaptivity into N-pair loss and contrastive loss, respectively.

To better qualitatively evaluate the learnt embedding representations, we further show the Barnes-Hut t-SNE \cite{van2014accelerating} visualizations of image embedding representations learnt by our DML-DA$_{con}$ on Cars196 dataset, CUB-200-2011 and Stanford Online Products datasets in Figure \ref{fig:figs_example_car}, \ref{fig:figs_example_bird} and \ref{fig:figs_example_product}, respectively. Specifically, we leverage all the images in the test split of each dataset and the 128-dimensional embedding representations of images are then projected into 2-dimensional space using Barnes-Hut t-SNE \cite{van2014accelerating}. It is clear that our learnt embedding representation effectively clusters semantically similar cars/birds/products despite of the significant variations in view point, pose and configuration.

\section{Conclusion}\label{sec:CON}
In this paper we have investigated the problem of training deep neural networks that are capable of high generalization performance in the context of metric learning. Particularly, we propose a new principle of density adaptivity into the learning of DML, which could lead to the largest possible intra-class variation in the embedding space. More importantly, the density adaptivity can be easily integrated into any existing DML implementations by simply adding one regularizer to the original objective loss. To verify our claim, we have strengthened three types of embedding, i.e., contrastive embedding, N-pair embedding and triplet embedding, with density regularizer. Extensive experiments conducted on three datasets validate our proposal and analysis. More remarkably, we achieve new state-of-the-art performance on all the three datasets. One possible future research direction would be to generalize our density adaptivity scheme to other types of embedding or other tasks with a large amount of classes.

\bibliographystyle{IEEEtran}
\bibliography{egbib}

\end{document}